\documentclass[10pt,twocolumn,letterpaper]{article}

\usepackage[pagenumbers]{cvpr} 
\usepackage{mmstyle}

\usepackage[dvipsnames]{xcolor}
\newcommand{\red}[1]{{\color{red}#1}}

\newcommand\blfootnote[1]{%
  \begingroup
  \renewcommand\thefootnote{}\footnote{#1}%
  \addtocounter{footnote}{-1}%
  \endgroup
}
\usepackage{colortbl}
\definecolor{shareblue}{RGB}{52, 102, 186}
\usepackage{graphicx}
\usepackage[normalem]{ulem}
\newcommand{\lin}[1]{\textcolor{red}{#1}}

\usepackage{pifont}
\newcommand{\cmark}{\ding{51}}%
\newcommand{\xmark}{\ding{55}}%

\usepackage[pagebackref,breaklinks,colorlinks]{hyperref}

\def\paperID{****} 
\def\confName{CVPR}
\def\confYear{2024}

\title{%
\raisebox{-1.1ex}{\protect\includegraphics[height=2.7\fontcharht\font`\B]{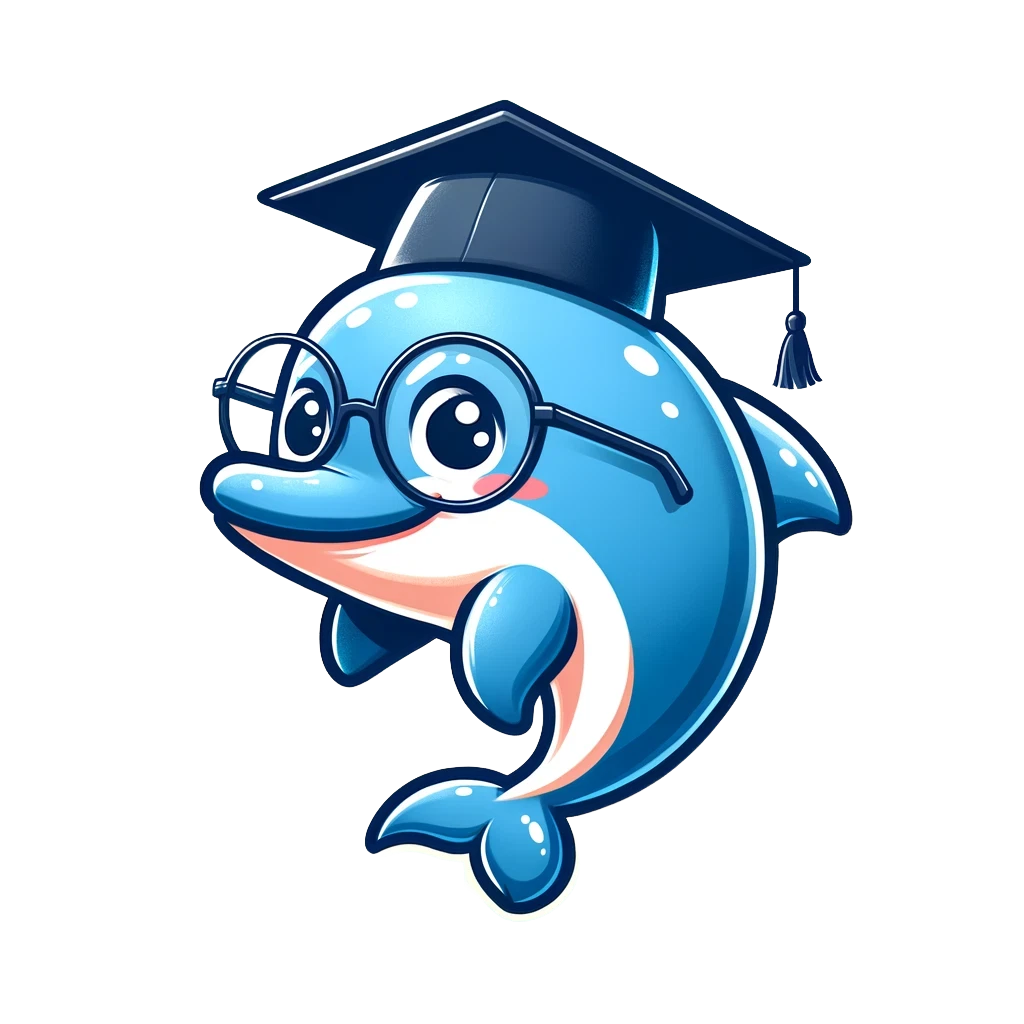}}
ShareGPT4V: Improving Large Multi-Modal Models with Better Captions}

\author{
Lin Chen$^{*\S1,2}$, Jinsong Li$^{*\S2}$, Xiaoyi Dong$^{2}$, Pan Zhang$^{2}$, Conghui He$^{2}$, Jiaqi Wang$^{2}$, \\ Feng Zhao$^{\dagger1}$, Dahua Lin$^{\dagger2}$\\
$^1$University of Science and Technology of China \quad 
$^2$Shanghai AI Laboratory \\
{\tt\small chlin@mail.ustc.edu.cn}, {\tt\small \{lijingsong, dongxiaoyi, zhangpan, heconghui, wangjiaqi\}.pjlab.org.cn} \\
{\tt\small fzhao956@ustc.edu.cn}, {\tt\small dhlin@ie.cuhk.edu.hk} 
}

\begin{document}

\twocolumn[{
\maketitle
\vspace{-8mm}
\begin{center}
    \centering
    \captionsetup{type=figure}
    \includegraphics[width=\textwidth]{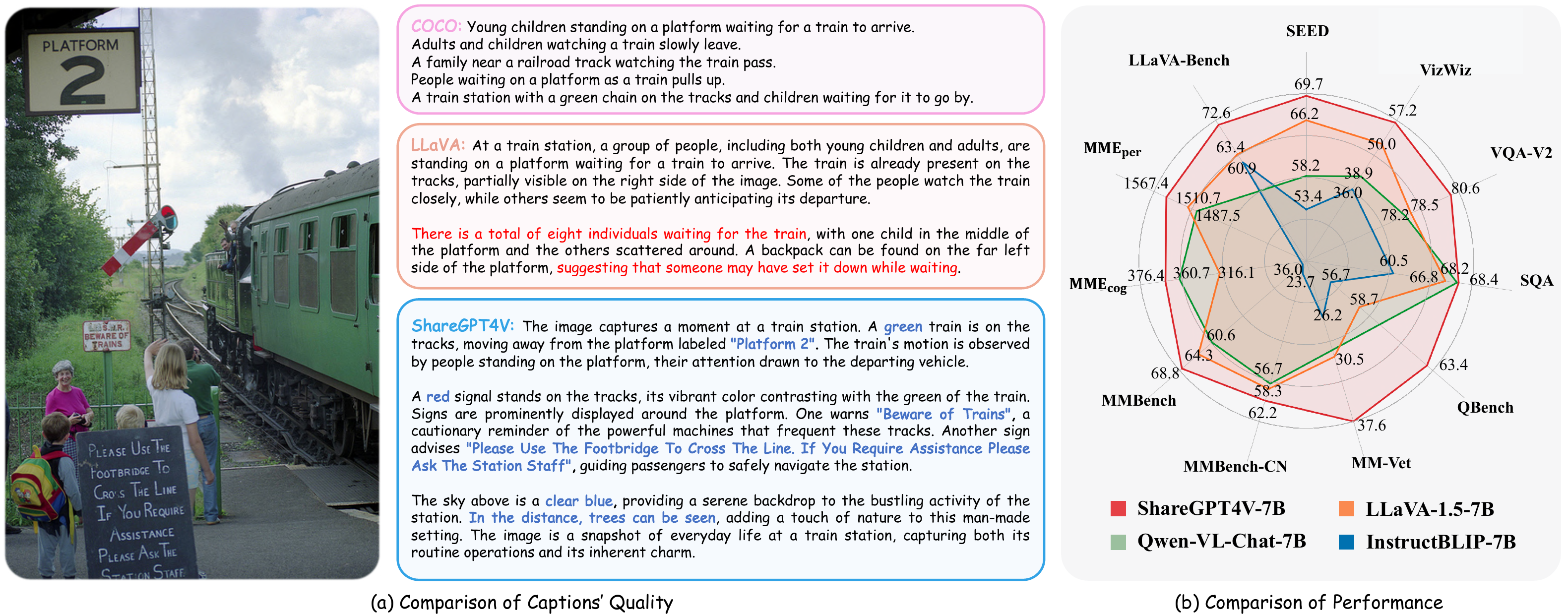}
    \vspace{-5mm}
    \captionof{figure}{
    (a) \textbf{We showcase a comparison between the caption} in our proposed ShareGPT4V dataset and those utilized by recent large multi-modal models (LMMs). Unlike COCO-Caption \cite{chen2015microsoft} involves brief human-made captions on the main subject. LLaVA-Instruct \cite{liu2023visual} combines human-made captions, bounding boxes, and GPT4 \cite{chatgpt} to `imagine' the image details, which leads to inevitable error/hallucination description (marked in \red{red}).  
    Our approach involves feeding carefully designed prompts along with images directly into the advanced GPT4-Vision \cite{gpt4v} and the descriptions are more detailed and accurate (marked in \textcolor{shareblue}{blue}). (b) \textbf{We highlight the remarkable performance} of the proposed LMM, ShareGPT4V-7B, developed with the assistance of the ShareGPT4V dataset. 
    }
    \label{fig:teaser}
\end{center}
}]

\blfootnote{$^*$ Equal contribution.\ \ $\dagger$ Corresponding authors. \ \ $\S$ Work done during an internship in Shanghai AI Laboratory.}
\begin{abstract}
In the realm of large multi-modal models (LMMs), efficient modality alignment is crucial yet often constrained by the scarcity of high-quality image-text data.
To address this bottleneck, we introduce the ShareGPT4V dataset, a pioneering large-scale resource featuring \textbf{1.2 million} highly descriptive captions, which surpasses existing datasets in diversity and information content, covering world knowledge, object properties, spatial relationships, and aesthetic evaluations.
Specifically, ShareGPT4V originates from a curated 100K high-quality captions collected from advanced GPT4-Vision and has been expanded to 1.2M with a superb caption model trained on this subset.
ShareGPT4V first demonstrates its effectiveness for the Supervised Fine-Tuning (SFT) phase, by substituting an equivalent quantity of detailed captions in existing SFT datasets with a subset of our high-quality captions, significantly enhancing the LMMs like LLaVA-7B, LLaVA-1.5-13B, and Qwen-VL-Chat-7B on the MME and MMBench benchmarks, with respective gains of \textbf{222.8/22.0/22.3} and \textbf{2.7/1.3/1.5}.
We further incorporate ShareGPT4V data into both the pre-training and SFT phases, obtaining ShareGPT4V-7B, a superior LMM based on a simple architecture that has remarkable performance across a majority of the multi-modal benchmarks.
This project is available at \url{https://ShareGPT4V.github.io} to serve as a pivotal resource for advancing the LMMs community.

\end{abstract}
\vspace{-3mm}    
\section{Introduction}
\label{sec:intro}

\begin{figure*}[t!]
\centering
\includegraphics[width=\textwidth]{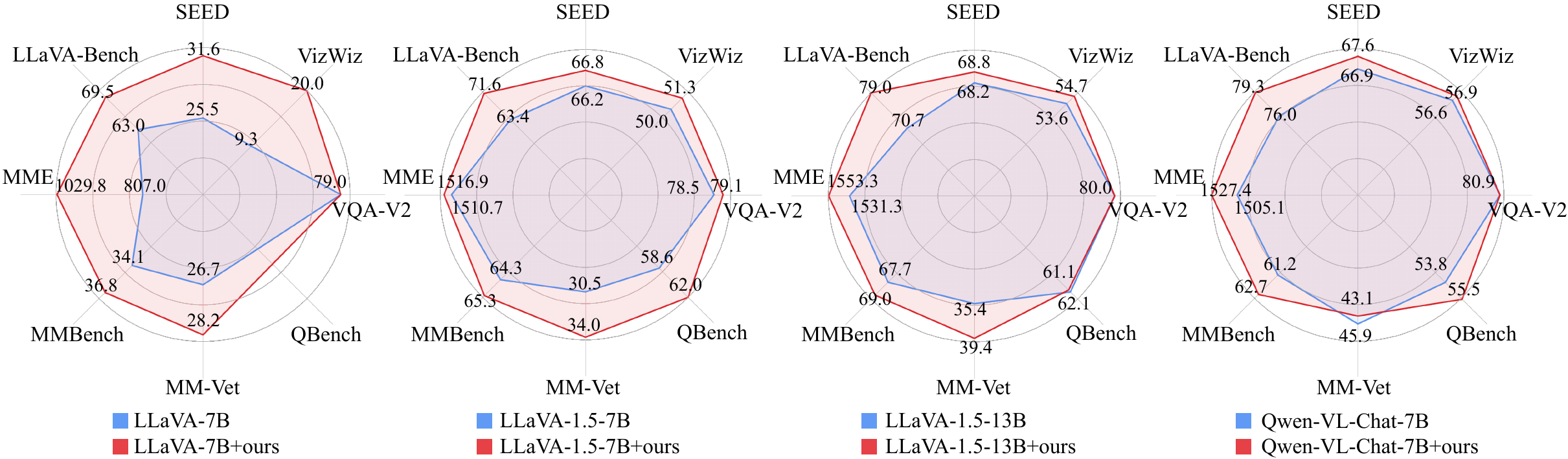}
\vspace{-3mm}
\caption{
\textbf{Illustration of the benefits high-quality captions bring to the SFT stage.} We compare the performance of various large multi-modal models before and after replacing a corresponding portion of their SFT captions with those generated by GPT4-Vision. The replacement ratio is only $3.5\%$ for LLaVA-1.5 \cite{liu2023improved} and Qwen-VL-Chat \cite{bai2023qwenvl}\protect\footnotemark, and $14.5\%$ for LLaVA \cite{liu2023visual}.
}
\vspace{-3mm}
\label{fig:equal_replace}
\end{figure*}

Recent breakthroughs in artificial intelligence have been driven notably by the development of large language models (LLMs) \cite{yang2023baichuan,brown2020language,chiang2023vicuna,chowdhery2022palm,du2021glm,bai2023qwen,touvron2023llama}.
Following the evolution, modality unification via LLMs becomes the inevitable tendency, and visual-aligned multi-modal LLMs\cite{zhu2023minigpt,chen2023minigpt,liu2023visual,liu2023improved,instructblip,zhang2023internlm,zhang2023llama,bai2023qwenvl,ye2023mplug,luo2023cheap} have witnessed ever-changing advances in recent days.
Putting aside the diversity in model architecture and training data, most of the large multi-modal models (LMMs) adhere to a dual-phase paradigm encompassing a pre-training stage with large-scale image-text pairs for modality alignment, followed by a supervised fine-tuning (SFT) stage that enhances multi-modal capabilities through instruction-format data.

Despite their efforts and achievements, we argue that the current LMMs still align the modalities in a sub-optimal manner, primarily due to the lack of sufficient high-quality image-text pairs. 
Vision, inherently rich in information and fine-grained semantics, is often reduced to simplistic captions in mainstream image-text datasets. 
These captions, typically brief and focused on salient objects, lead to a significant reduction in information content and sub-optimal modality alignment.

\footnotetext{The pre-SFT checkpoint and corresponding SFT data of Qwen-VL-Chat-7B is not public avilable. So we fine-tuned its final checkpoint using the 665K SFT dataset collected by LLaVA-1.5 and reported these results for a fair and controllable comparison.}

To prove our argument, we conducted a straightforward experiment: we substituted the image-text pairs utilized in the SFT stage of several typical LMMs with equivalent comprehensive captions generated by the advanced GPT4-Vision model and re-benchmarked these LMMs. As shown in Figure \ref{fig:equal_replace}, such equivalent substitution, despite its relatively minimal extent (only 3.5\% of the SFT data in the LLaVA-1.5 case), resulted in consistent performance gains across various LMMs and benchmarks.
Encouraged by these promising results, we expanded our efforts to collect high-quality captions on a larger scale, involving two phases. In the initial phase, approximately 100K images from various data sources were gathered. We employed carefully designed data-specific prompts to effectively utilize GPT4-Vision to generate high-quality descriptions. The resulting captions, averaging \textbf{942 characters}, encompass a comprehensive range of image information, such as world knowledge, object properties, spatial relation, aesthetic evaluation, etc. In the second phase, we utilize these captions to build a strong caption model, which gets rid of the data source specialized prompt and could generate comprehensive captions for given images.

Based on the above endeavors, we introduce the ShareGPT4V dataset, the first highly descriptive image-text collection. It comprises two components: 100K GPT4-Vision generated captions with diverse image sources and 1.2M captions crafted by our caption model, which is learned from the 100K high-quality captions.
With the aid of this dataset, we have developed an eponymous state-of-the-art large multi-modal model, the ShareGPT4V-7B. To maintain clarity in our discourse, `dataset' or `model' will be distinctly specified when referring to ShareGPT4V. 
Figure \ref{fig:teaser}(b) shows that ShareGPT4V-7B outperforms other advanced 7B-scale LMMs in all 11 benchmarks, showcasing its competitive performance.
For instance, our ShareGPT4V-7B model achieves an impressive total score of 1943.8 on the MME benchmark, surpassing the second-ranked Qwen-VL-Chat-7B model, which was trained on 1.4 billion samples, by 95.6 points.

In a nutshell, our contributions are threefold:
\begin{itemize}[leftmargin=*]
\setlength\itemsep{.2em}
\item We point out the fact that existing low-quality captions can impede the alignment between vision and language modalities of LMMs and we verify it with experimental results. 
This revelation highlights an urgent requirement within the LMM community for high-quality captions to effectively alleviate such a dilemma.

\item  We introduce the ShareGPT4V dataset, a large-scale image-text collection featuring 100K highly descriptive captions generated by GPT4-Vision and 1.2M high-quality captions generated by our caption model. The caption covers world knowledge, object attributes, spatial relations, aesthetic assessment, etc. Moreover, the general caption model trained on entire GPT4-Vision-generated captions could further scale our dataset and will also be available for community usage.

\item Leveraging the proposed dataset, we have developed the ShareGPT4V-7B, an advanced large multimodal model. Despite without elaborate architecture design, this model consistently demonstrates impressive performance across various multi-modal benchmarks.
\end{itemize}

\section{Related Work}
\label{sec:related}

\textbf{Large Language Models.} In recent years, with the surge in data and computational power, the development of large language models has experienced a boom. Early encoder-decoder models like BERT \cite{devlin2018bert} and T5 \cite{raffel2020exploring}, and decoder-centric models such as GPT \cite{radford2018improving}, leveraged the Transformer architecture \cite{vaswani2017attention} to excel in various NLP tasks.  
The success in GPT3 \cite{brown2020language} has popularized the use of decoder-only architectures, which rely on auto-regressive decoding for generating predictions.
Subsequent models like PaLM \cite{chowdhery2022palm} extended the limits of model parameters and dataset scale, while others like InstructGPT \cite{ouyang2022training} and ChatGPT \cite{chatgpt} introduced fine-tuning and reinforcement learning techniques for improved conversational interaction. These developments, along with contributions from the open-source community \cite{chiang2023vicuna,touvron2023llama,touvron2023llama,yang2023baichuan,team2023internlm}, have set new benchmarks and opened avenues for future research in NLP area.

\noindent{\textbf{Large Multi-modal Models.}} As LLMs rapidly evolve, a faction within the research community is increasingly concentrating on introducing visual knowledge into LLMs. 
Central to this area are the seminal works in modality alignment within the vision-language learning area \cite{radford2021learning,jia2021scaling}. A notable instance is CLIP \cite{radford2021learning}, which exemplifies the alignment of visual and textual modalities through contrastive learning on extensive image-text pairs. A series of works \cite{li2022blip,li2023blip} were improved upon CLIP by employing refined data strategies for more diverse data, they have been effective for basic visual tasks \cite{li2022grounded,liu2023grounding,zhang2022glipv2} but less so for complex tasks like visual question answering.
MiniGPT-4 \cite{chen2023minigpt}, leveraging an LLM \cite{chiang2023vicuna} and a visual encoder \cite{fang2023eva}, has shown proficiency in image-text dialogues through pre-training alignment and instruction fine-tuning. Subsequent research \cite{liu2023visual,li2023otter,instructblip,ye2023mplug,chen2023shikra,bai2023qwenvl,peng2023kosmos} has further enhanced LMMs by focusing on the quality and diversity of pretraining and fine-tuning data. For instance, LLaVA \cite{liu2023visual} and InstructBLIP \cite{instructblip}, with improved instruction fine-tuning, have advanced the understanding of complex prompts. mPLUG-Owl \cite{ye2023mplug}, Shikra \cite{chen2023shikra}, and KOSMOS-2 \cite{peng2023kosmos} have introduced new data types and training techniques, like grounding data, to reduce hallucinations and improve LMMs' grounding capability. Regrettably, it appears that the current LMMs have somewhat overlooked a crucial element: the quality of captions in image-text pairs.

\noindent{\textbf{Image-text Data Enhancement.}} In the vision-language learning area, several initiatives \cite{fan2023improving,gadre2023datacomp,nguyen2023improving,lai2023scarcity} have been undertaken to enhance the quality of captions within image-text pairs. LaCLIP \cite{fan2023improving} leverages LLMs to rewrite raw captions, but its effectiveness is often hindered by hallucinations due to limited visual information and the low quality of original captions. Research \cite{gadre2023datacomp,nguyen2023improving} explores methods to filter and blend raw and synthetic captions to enhance the CLIP model. A recent work, VeCLIP \cite{lai2023scarcity}, proposes using LLMs to amalgamate information from both raw and synthetic captions. Nevertheless, the approach is constrained by the low quality of synthetic captions, resulting in only minimal incorporation of visual knowledge in the caption fusion process. To the best of our knowledge, in the LMM area, LLaVA \cite{liu2023visual} uniquely inputs human-annotated short captions and bounding boxes into the GPT4 language model. This approach lets the model `imagine' viewing the image before producing detailed captions. However, this method relies heavily on extensive human-annotated data and does not allow the model to truly `see' the images. Consequently, it tends to generate detailed descriptions primarily of main objects, often including those in obscure corners but annotated with bounding boxes, leading to potential hallucinations in the LMMs' output. In contrast, we employ the most advanced LMM, GPT4-Vision, which is capable of directly producing highly descriptive captions from deliberated prompts and corresponding image inputs.

\begin{figure*}[t!]
\begin{center}
\includegraphics[width=\textwidth]{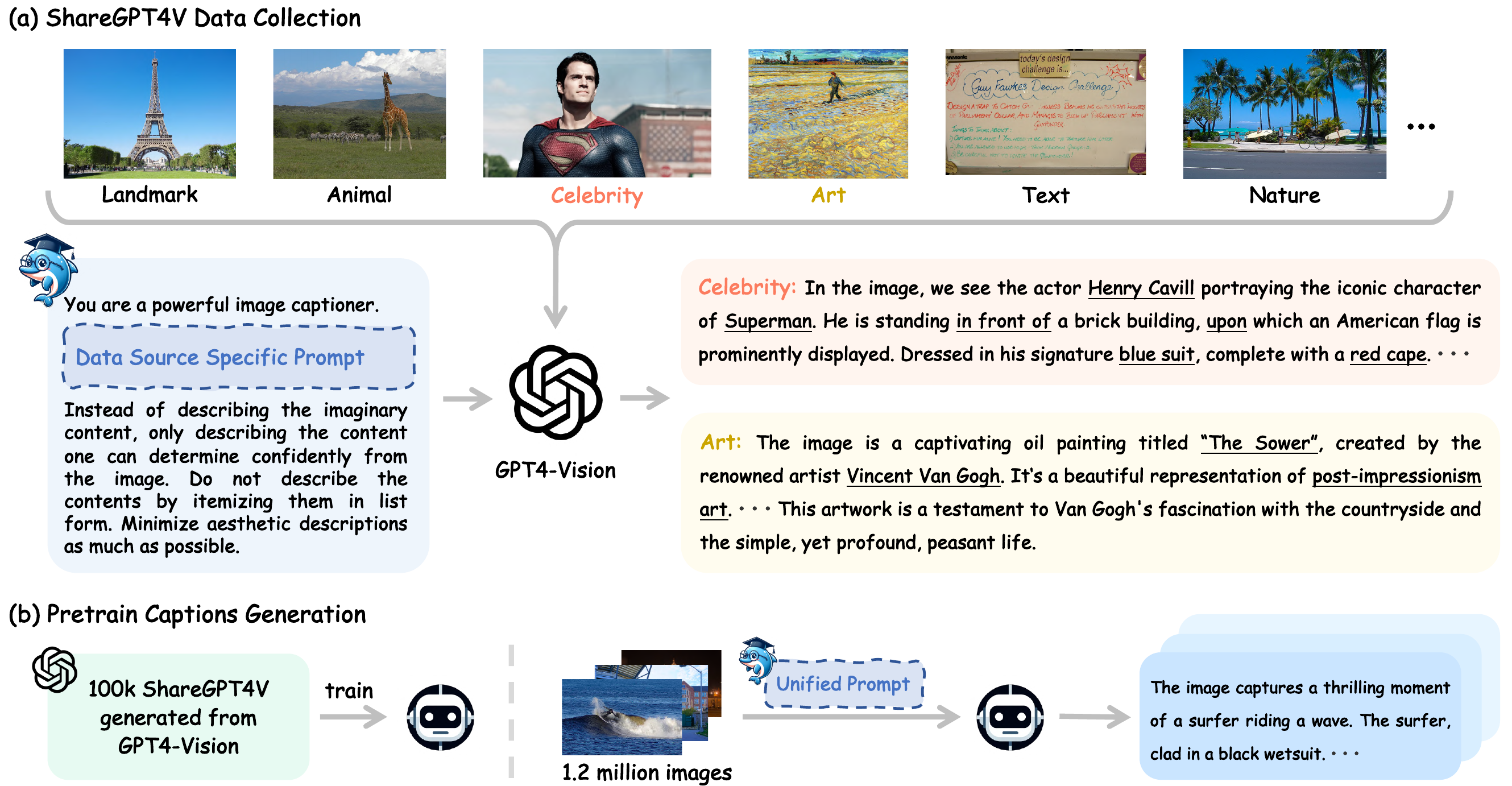}
\end{center}
\vspace{-2mm}
\caption{
\textbf{An overview for crafting the ShareGPT4V dataset.} (a) We illustrate the procedure for collecting highly descriptive captions from GPT4-Vision \cite{gpt4v} via various image sources and data-specific prompts, resulting in 100K high-quality captions that encapsulate a wide array of information conveyed by the images. (b) We delineate the process of utilizing the seed captions to train a general captioner and then employing this captioner to generate 1.2M high-quality captions for pre-training usage.
}
\vspace{-2mm}
\label{fig:data_source}
\end{figure*}

\section{ShareGPT4V Dataset}
\label{sec:sharegpt4v_dataset}

\subsection{Overview}
In this section, we provide a detailed exposition of the process involved in creating the ShareGPT4V dataset. 
Subsection \ref{sec:sft_data} elaborates on how we utilized GPT4-Vision to generate 100K high-quality captions from various image sources and briefly validates their significant role in the SFT phase of LMMs. 
Subsection \ref{sec:pt_data} describes our methodology for reasonably expanding the 100K high-quality captions in Sec.\ref{sec:sft_data} to 1.2M captions, matching the quality generated by GPT4-Vision with acceptable cost. Table \ref{tav:data_stat} presents a comparison between our dataset and existing widely-used caption datasets in the LMM field. Our ShareGPT4V dataset stands out due to its more diverse range of image sources, the use of a more advanced caption producer, a larger number of samples, and the generation of longer captions.

\subsection{ShareGPT4V Data Collection}
\label{sec:sft_data}
The supervised fine-tuning captions were collected from GPT4-Vision, the latest and most advanced LMM. For each image selected from a specific data source $D$, we employed a meticulously crafted, data-specific prompt $P_{D}$. This prompt instructed GPT4-Vision to generate detailed descriptions, taking into account factors such as world knowledge, object attributes, spatial relationships, and aesthetic evaluations.

\begin{table}[h!]
\centering
\footnotesize
\resizebox{1.0\linewidth}{!}{
\setlength{\tabcolsep}{0.5mm}{
\begin{tabular}{l|lclll}
\toprule
Name & Image Source &Visible & Captioned by  & Samples & Avg. \\ \midrule
COCO-Caption \cite{chen2015microsoft} & COCO \cite{lin2014microsoft} &\checkmark & Human & 118K & 52 \\
BLIP-LCS \cite{li2022blip} & LCS &\checkmark & BLIP \cite{li2022blip} & 558K & 54 \\
LLaVA-23K \cite{liu2023visual} & COCO \cite{lin2014microsoft} &$\times$ & GPT4 \cite{chatgpt} & 23K & 609 \\ \midrule
ShareGPT4V & LCS, COCO \cite{lin2014microsoft}, etc &\checkmark & GPT4-Vision \cite{gpt4v} & 100K & \textbf{942} \\
ShareGPT4V-PT & LCS, COCO \cite{lin2014microsoft}, etc &\checkmark & Share-Captioner & \textbf{1,246K} & 826 \\ \bottomrule
\end{tabular}
}}
\vspace{-2mm}
\caption{
\textbf{Comparison of widely-used caption datasets and ShareGPT4V.} `LCS' abbreviates the LAION \cite{schuhmann2021laion}, CC \cite{sharma2018conceptual}, and SBU \cite{saleh2015large} datasets. The `Visible' column denotes the image visibility during captioning, and the last column shows the average character number of the caption.
}\label{tav:data_stat}
\vspace{-4mm}
\end{table}
\noindent\textbf{Data sources}. 
To maximize the diversity and comprehensiveness of our data, we compiled around 100K images from various data sources, including images for detection~\cite{lin2014microsoft} and segmentation~\cite{kirillov2023segment}, complex text-containing images\cite{sidorov2020textcaps}, as well as various web images~\cite{schuhmann2021laion,sharma2018conceptual,ordonez2011im2text} containing artworks, landmarks, celebrities \etc. More details could be found in the supplementary material.

\begin{figure*}[t!]
\centering
\includegraphics[width=1\textwidth]{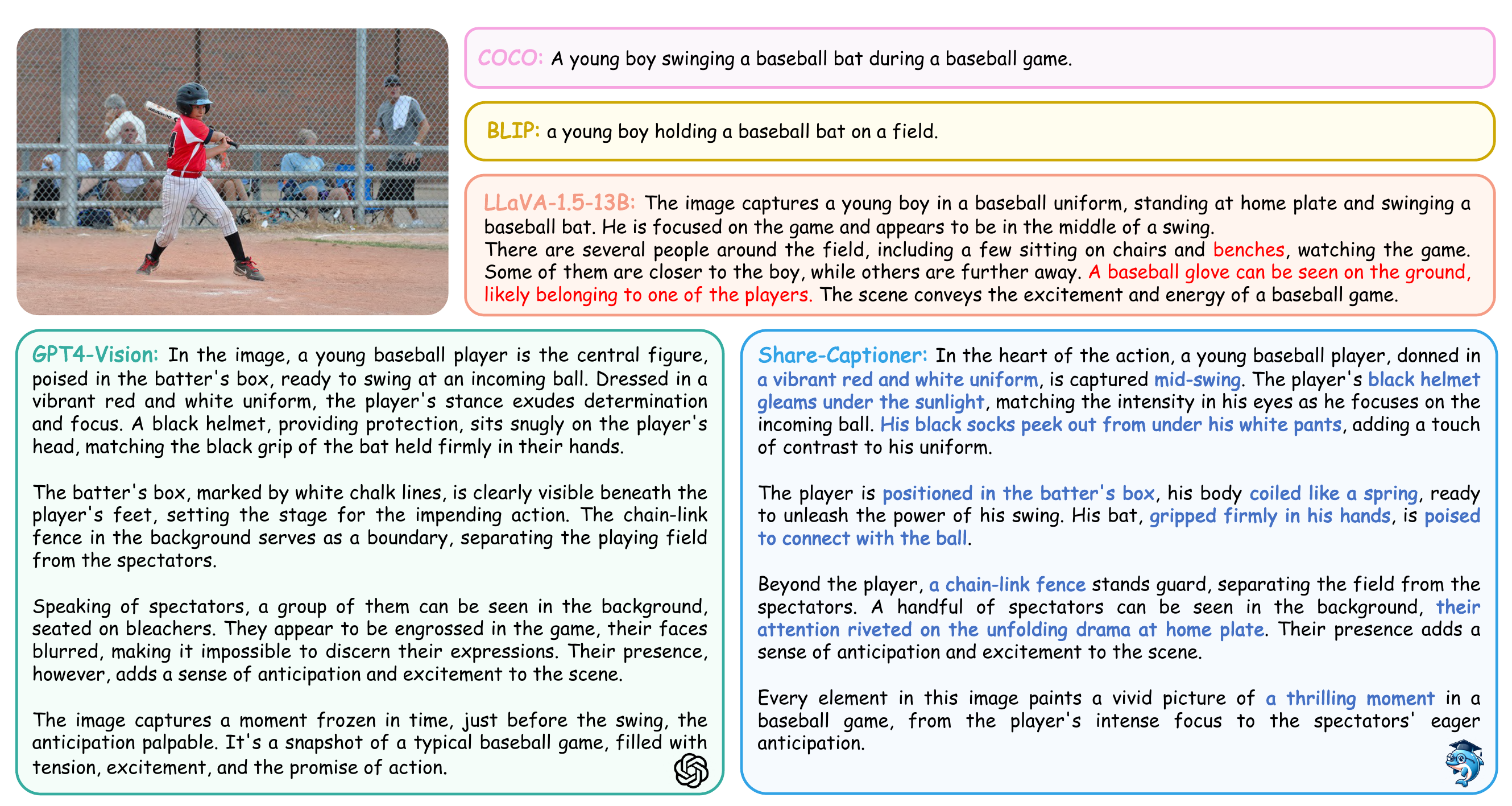}
\caption{
\textbf{A qualitative comparison of caption quality from various sources.} The COCO \cite{chen2015microsoft} captions were generated by humans and the remaining captions were produced by large multi-modal models using the same prompts and images. Mistakes within the captions are highlighted in \lin{red}, whereas detailed and accurate parts are emphasized in \textcolor{shareblue}{blue}. Notably, the image used for this comparison was not included in the training set of our Share-Captioner.
}
\vspace{-4mm}
\label{fig:cap_comp}
\end{figure*}

\noindent\textbf{Prompt Design}.
Given the diversity of our image sources, we expect a highly content-related description for each image. That is, the captions should extend beyond mere appearance and attributes, incorporating knowledge-related information.
For instance, the Eiffel Tower should not be simply described as a tall iron tower, and a picture of Einstein should not be concluded as an old man.

For the description quality and stability, we designed a base prompt for a general description and added a specialized prompt for each data source. 
The base prompt asks the GPT4-Vision to describe the basic information of the image, including the object attributes, appearance, and spatial relationships. 
The specialized prompt focuses on some data-related information, as shown in Figure \ref{fig:data_source}, we emphasize that the GPT4-Vision should mention some corresponding knowledge, such as the name and geographical location of a landmark-related image.
Additionally, we add an aesthetic-related prompt for part of the images, to further improve the comprehensiveness of the description.

\noindent\textbf{Quality Verification}.
We conducted a straightforward experiment to verify the quality of the collected data:
we chose a range of advanced, publicly available LMMs, including LLaVA-7B \cite{liu2023visual}, LLaVA-1.5-7B \cite{liu2023improved}, LLaVA-1.5-13B \cite{liu2023improved}, and Qwen-VL-Chat-7B \cite{bai2023qwenvl}. For a fair comparison, we replaced a corresponding portion of detailed captions in their Supervised Fine-Tuning (SFT) datasets with a selection from our 100K GPT4-Vision-generated captions, while maintaining image data sources as consistent as possible. As depicted in Figure \ref{fig:equal_replace}, the integration of our highly descriptive captions significantly improved the SFT phase performance across these varied LMMs, reinforcing our pursuit to gather more high-quality captions for potential benefits in the pretraining stage.

\subsection{ShareGPT4V-PT Data Generation}
\label{sec:pt_data}
Compared with the supervised fine-tuning stage, modality alignment in the pre-training phase is more crucial and demands an large-scale dataset. For building a pre-training dataset, we employed the 100K high-quality captions generated by GPT4-Vision to fine-tune an alternative caption model and we have named it as Share-Captioner.  
Thanks to its training on diverse and comprehensive data, the Share-Captioner is capable of generating highly content-related descriptions with unified instruction. This approach allows the data scaling phase to proceed without the need for specialized prompt design. 

To amass a substantial volume of high-quality image-text pairs, we selected a subset of 1.2 million images from current public datasets (see supplementary material for more details) and employed our pre-trained Share-Captioner for the captioning process.
The entire caption generation process required around 44 A100 GPU days and we name this part of data as ShareGPT4V-PT.

\begin{table}[t]
\centering
\resizebox{1\linewidth}{!}{
\setlength{\tabcolsep}{1mm}{
\begin{tabular}{l|ccc}
\toprule
Preference & GPT4-Vision & Share-Captioner & Comparable \\ \midrule
Percentage &38.2$\%$  &35.3$\%$  &26.5$\%$  \\ \bottomrule
\end{tabular}
}}
\vspace{-2mm}
\caption{\textbf{Human evaluation} on Share-Captioner vs. GPT4-Vision over 100 validation samples and 10 volunteers.} \label{table:user_study}
\vspace{-4mm}
\end{table}
\noindent\textbf{Qualitative Analysis}. 
For qualitative analysis, Figure \ref{fig:cap_comp} presents caption results from human-made COCO-Captions \cite{chen2015microsoft}, BLIP \cite{li2022blip}, LLaVA-1.5-7B \cite{liu2023improved}, Share-Captioner, and GPT4-Vision. It is important to note that the images featured in this figure were not part of the training dataset for Share-Captioner. The results depicted in Figure \ref{fig:cap_comp} demonstrate that Share-Captioner produced results that are closely comparable to those generated by GPT4-Vision, aligning with our anticipated capabilities for the captioning process.

\noindent\textbf{Quantitative Analysis}. 
As detailed in Table \ref{table:user_study}, we generate 100 captions with
GPT4-Vision and our Share-Captioner, and invite 10 volunteers to select the better one. As anticipated, our Share-Captioner performs on par with the GPT4-Vision, confirming the quality of the ShareGPT4V dataset.

\begin{table*}[t!]
\centering
\resizebox{0.99\textwidth}{!}{
\begin{tabular}{ll|lllllllllll}
\toprule
Method & Language Model & LLaVA$^{W}$ & MME$^{P}$ & MME$^{C}$ & MMB & MMB$^{CN}$ & SEED$^{I}$ & MM-Vet & QBench & SQA$^{I}$ & VQA$^{V2}$ & VizWiz \\ \midrule
BLIP-2 & FLAN-T5 & 38.1 & 1293.8 & 290.0 & - & - & 46.4 & 22.4 & - & 61.0 & 41.0 & 19.6 \\
InstructBLIP & Vicuna-7B & 60.9 & - & - & 36.0 & 23.7 & 53.4 & 26.2 & 56.7 & 60.5 & - & 34.5 \\
InstructBLIP & FLAN-T5 & 58.2 & 1212.8 & 291.8 & - & - & - & 25.6 & - & 63.1 & - & 33.4 \\
Shikra & Vicuna-13B & - & - & - & 58.8 & - & - & - & 54.7 & - & 77.4 & - \\
IDEFICS-80B & LLaMA-65B & - & - & - & 54.5 & 38.1 & - & - & - & - & 60.0 & 36.0 \\
Qwen-VL & Qwen-7B & - & - & - & 38.2 & 7.4 & 56.3 & - & 59.4 & 67.1 & 78.8 & 35.2 \\
Qwen-VL-Chat & Qwen-7B & - & 1487.5 & \underline{360.7} & 60.6 & 56.7 & 58.2 & - & - & 68.2 & 78.2 & 38.9 \\
LLaVA & Vicuna-7B & 63.0* & 807.0* & 247.9* & 34.1* & 14.1* & 25.5* & 26.7* &-  & 38.5* & 79.0* & 9.3* \\
LLaVA-1.5 & Vicuna-7B & 63.4 & 1510.7 & 316.1* & 64.3 & 58.3 & 66.2* & 30.5 & 58.7 & 66.8 & 78.5 & 50.0 \\
LLaVA-1.5 & Vicuna-13B & \underline{70.7} & \underline{1531.3} & 295.4* & \underline{67.7} & \textbf{63.6} & \underline{68.2} & \underline{35.4} & \underline{62.1} & \textbf{71.6} & \underline{80.0} & \underline{53.6} \\ \midrule
\rowcolor[HTML]{F2F3F5} 
ShareGPT4V-7B & Vicuna-7B & \textbf{72.6} & \textbf{1567.4} & \textbf{376.4} & \textbf{68.8} & \underline{62.2} & \textbf{69.7} & \textbf{37.6} & \textbf{63.4} & \underline{68.4} & \textbf{80.6} & \textbf{57.2} \\ \bottomrule
\end{tabular}
}
\vspace{-2mm}
\caption{\textbf{Comparison with SoTA methods on 11 benchmarks.} With 7B parameters, ShareGPT4V-7B outperforms competitors in 9 out of 11 benchmarks and ranks second on the others, despite these competitors using larger training datasets or more parameters. Benchmark names are abbreviated due to space limits. LLaVA$^{W}$: LLaVA-Bench (In-the-Wild) \cite{liu2023visual}; MME$^{P}$: MME Perception \cite{fu2023mme}; MME$^{C}$: MME Cognition \cite{fu2023mme}; MMB: MMBenchmark \cite{liu2023mmbench}; MMB$^{CN}$: MMBench-Chinese \cite{liu2023mmbench}; SEED$^{I}$: SEED-Bench (Image) \cite{li2023seed}; MM-Vet \cite{yu2023mm}; QBench \cite{wu2023q}; SQA$^{I}$: ScienceQA-IMG \cite{lu2022learn}; VQA$^{V2}$ \cite{goyal2017making}; VizWiz \cite{gurari2018vizwiz}. * indicates our re-implemented test results missed in benchmarks or origin papers. The best results are \textbf{bold} and the second-best results are \underline{underlined}.}\label{tab:entire_comp}
\vspace{-4mm}
\end{table*}
\vspace{-2mm}
 
\section{ShareGPT4V-7B Model}
\label{sec:method}
To ascertain the efficacy of the ShareGPT4V dataset, we conducted experiments within a fair and controlled setting. This led to the development of ShareGPT4V-7B, a streamlined yet superior baseline LMM leveraging the high-quality data from the ShareGPT4V dataset in both the pre-training and SFT stages.

\subsection{Model Architecture}
The ShareGPT4V-7B model follows the design of LLaVA-1.5 \cite{liu2023improved}, including three integral components: (1) A vision encoder utilizing the CLIP-Large model \cite{radford2021learning}, with a resolution of 336$\times$336 and a patch size of 14, converting input images into 576 tokens. (2) A projector, which is a two-layer multi-layer perception (MLP), is introduced to connect the vision and language modalities. (3) A LLM, based on the open-source Vicuna-v1.5 \cite{chiang2023vicuna}, derived from LLaMA2 \cite{touvron2023llama}. Currently, our focus is on the lightweight 7B model scale, and we have empirically validated that even with lightweight training data and model scale, it can significantly outperform many current LMMs that utilize extensive training datasets or larger model scales.

\vspace{2mm}

\subsection{Pre-Training}
In the pre-training stage, we utilize the pre-training subset of the ShareGPT4V dataset, i.e., ShareGPT4V-PT. Given these high-quality captions, solely fine-tuning the MLP does not suffice to exploit their full capabilities. In previous LMM research \cite{liu2023visual,liu2023improved,zhu2023minigpt,chen2023minigpt}, the vision encoder is generally not fine-tuned during pre-training, a rational approach considering the lower quality of previously used captions, where fine-tuned the vision encoder might degrade its visual knowledge extraction ability. We opted for simultaneous fine-tuning of the vision encoder, projector, and large language model. With this configuration, the large language model acquires a native understanding of visual embeddings, while also prompting the vision encoder to create relevant visual embeddings for elements in captions. This setup enables a comprehensive exploration and understanding of the knowledge embedded in visual embeddings, aligned with the intricate details of the captions. Specifically, we consistently applied a learning rate of $2e^{-5}$ across all components, with a batch size set at $256$, and the comprehensive optimization process spanned roughly $4700$ steps. Notably, we experimentally found that selectively fine-tuning only the latter half of the vision encoder's layers achieves optimal results, coupled with a satisfactory level of training efficiency.

\begin{figure*}[t!]
\centering
\includegraphics[width=1\textwidth]{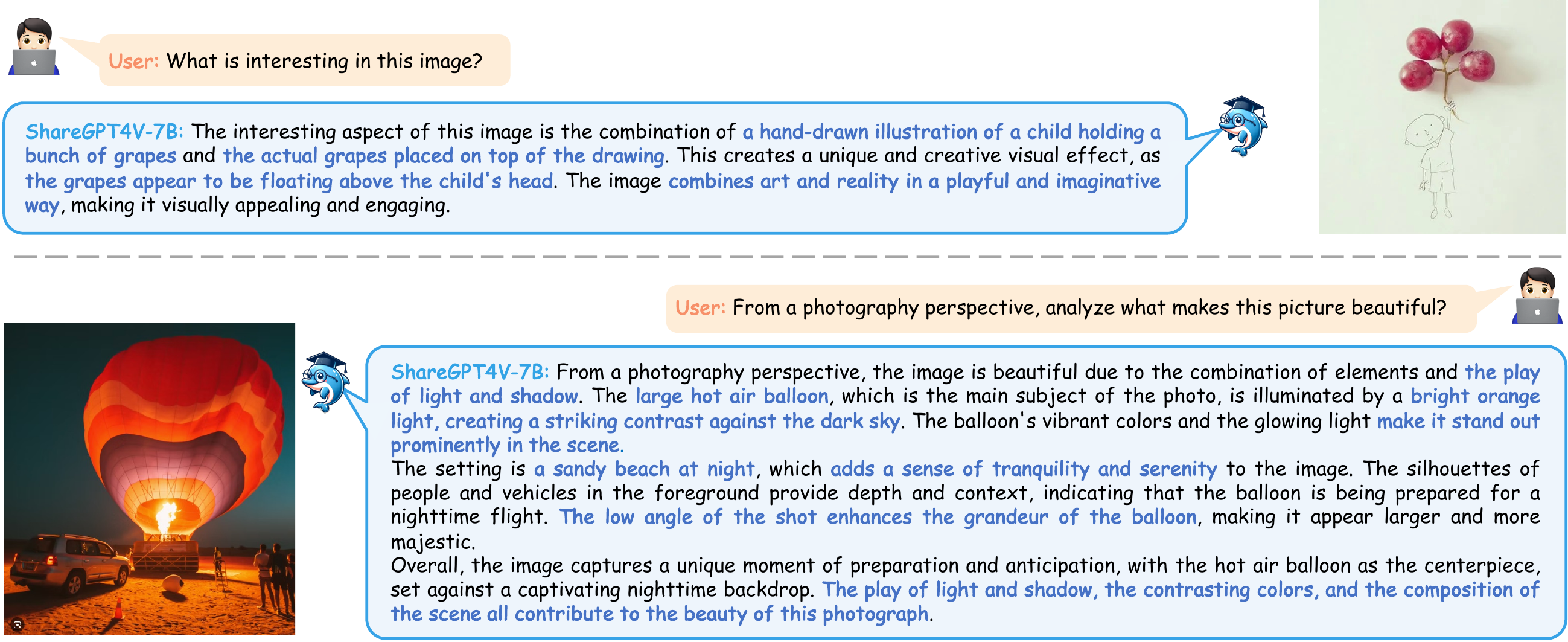}
\caption{
\textbf{Examples of multi-modal dialogue with ShareGPT4V-7B model.} High-quality content is highlighted in \textcolor{shareblue}{blue} for clarity.
}
\vspace{-3mm}
\label{fig:chat_case}
\end{figure*}

\vspace{2mm}
\subsection{Supervised Fine-Tuning.}
As we emphasized above, the goal of this paper is not to build a new SOTA model with some unique architecture designs but to \textbf{investigate the effectiveness of high-quality captions to realize better modality alignment of LMMs}.
So we utilize the 665k supervised data organized by LLaVA-1.5 and only replace part of it with our ShareGPT4V dataset. 
In detail, the 665k data is gathered from publicly available academic task-oriented data \cite{marino2019ok,schwenk2022okvqa,mishra2019ocr,sidorov2020textcaps,krishna2017visual,kazemzadeh2014referitgame,sharegpt} and instruction-tuning data for conversational and complex reasoning tasks \cite{liu2023visual} involving natural images \cite{lin2014microsoft}. 
It contains 23k detailed description data and we replaced it with randomly sampled 23K high-quality captions from the 100K captions in ShareGPT4V. 
During the SFT stage, to enhance the training efficiency and compare fairly, we froze the vision encoder and instead focused on fine-tuning the projector and the large language model. The learning rate was established at $2e^{-5}$, with a batch size of $128$, and the total optimization process spanned around $5200$ steps.
 
\section{Experiments}
\label{sec:exp}

\subsection{Benchmarks}
To thoroughly assess our proposed ShareGPT4V-7B model, we evaluate it across 11 benchmarks, covering a range of academic Visual Question Answering (VQA) tasks and recent benchmarks designed specifically for large multi-modal models (LMMs). The LLaVA (in the wild) benchmark \cite{liu2023visual} is composed of 60 questions, spanning three distinct tasks: conversation, complex reasoning, and detailed description. The MME Benchmark \cite{fu2023mme} evaluates LMMs' perception and cognition capabilities through a series of carefully crafted questions across 14 sub-tasks. MMBench and MMBench-CN \cite{liu2023mmbench} benchmarks manually design questions to evaluate the model's vision-related reasoning and perception abilities for English and Chinese, respectively. SEED \cite{li2023seed}, with the assistance of GPT4, generated a dataset comprising approximately 19K questions related to images and videos. MM-Vet \cite{yu2023mm} uses GPT4 for a six-dimensional LMM capability assessment. Q-Bench \cite{wu2023q} assesses low-level perception, while VQA-v2 \cite{goyal2017making} and VisWiz \cite{gurari2018vizwiz} are benchmarks in the realm of traditional Visual Question Answering (VQA) tasks.

\subsection{Quantitative Comparison}
We present a quantitative comparison between our proposed ShareGPT4V-7B model with existing state-of-the-art LMMs. Notably, compared with previous LMMs, our ShareGPT4V-7B attained the most superior performance in 9 out of the total 11 benchmarks. 

Specifically, our ShareGPT4V-7B model outperformed the previously best-performing LLaVA-1.5-13B model by 1.9 points on the LLaVA (in the wild) benchmark, demonstrating superior capabilities in tasks such as detailed description and complex reasoning. On the MME Benchmark, it achieved the highest scores in both perception (P) and cognition (C) capabilities, surpassing LLaVA-1.5-13B in perception by 36.1 points and exceeding Qwen-VL-Chat, which was trained on 1.4 billion data, by 15.7 points in cognition. Our model also achieved an optimal accuracy of 68.8$\%$ on MMBench, leading the second-best by 1.1$\%$. Furthermore, on the SEED (image) benchmark, which includes 9 assessment dimensions and 14K questions, ShareGPT4V-7B achieved the highest score of 69.7$\%$, 1.5$\%$ higher than the second-ranked LLaVA-1.5-13B. In the low-level image assessment QBench, our model's top score of 63.4$\%$ can be attributed to the diversity of our constructed dataset. Lastly, our model almost consistently performed best on traditional VQA benchmarks with the smallest model size. 

Our findings demonstrate to the community that even with a simple architecture, public data, and lighter parameters (7B), it is possible to outperform many competitors with massive training data and parameter sizes, thanks to the support of these high-quality captions.

\subsection{Multi-modal Dialogue}
In Figure \ref{fig:chat_case}, we present two representative examples within multi-modal dialogue scenarios. The figure demonstrates that our ShareGPT4V-7B exhibits satisfactory capabilities in understanding image details and performing aesthetic assessments. This further corroborates the significance of the high-quality captions we have collected.

\begin{table}[t]
\centering
\small
\vspace{-1mm}
\setlength{\tabcolsep}{1.5mm}{
\begin{tabular}{cc|ccc}
\toprule
\begin{tabular}[c]{@{}l@{}}Pre-training with  \\ ShareGPT4V-PT\end{tabular} & \begin{tabular}[c]{@{}l@{}}SFT with  \\ ShareGPT4V\end{tabular} & MME$^{P}$ & MMB & SEED$^{I}$ \\ \midrule
\xmark  & \xmark  & 1510.7 & 64.3 & 66.2 \\
\xmark & \cmark & 1542.1 & 66.8 & 66.7 \\
\cmark & \xmark  & 1557.2 & 67.4 & 68.5 \\
\cmark & \cmark & \textbf{1567.4} & \textbf{68.8} & \textbf{69.7} \\ \bottomrule
\end{tabular}
}
\vspace{-2mm}
\caption{\textbf{Ablation study of the training strategy.} The ShareGPT4V dataset improves the model performance in both the pre-training and supervised fine-tuning stages.}
\label{tab:ablation_component}
\vspace{-4mm}
\end{table}

\subsection{Ablations}
\noindent{\textbf{Effectiveness of ShareGPT4V Dataset.}} 
As shown in Table \ref{tab:ablation_component}, we conducted a thorough ablation study to assess the impact of the ShareGPT4V-PT and ShareGPT4V subsets. Our baseline is the LLaVA-1.5-7B model, without utilizing the ShareGPT4V dataset in either pretraining or SFT stages. Utilizing only our ShareGPT4V subset during the SFT stages resulted in a significant increase of 31.4 points in MME perception score, and improvements of 2.5$\%$ and 0.5$\%$ in accuracy on the MMBench and SEED benchmarks, respectively. Notably, ShareGPT4V used here was selected from various data sources, yielding more performance gains than those from solely the COCO dataset (see in Figure \ref{fig:equal_replace}). When only the ShareGPT4V-PT subset was used during pretraining, we observed a remarkable gain of 46.5 points in MME perception, along with substantial accuracy improvements of 3.1$\%$ and 2.3$\%$ on the MMBench and SEED benchmarks, respectively. Moreover, employing the ShareGPT4V dataset in both pretraining and SFT phases led to further satisfactory enhancements in overall performance, effectively validating the necessity of incorporating high-quality captions in both training stages.

\begin{table}[t]
\centering

\small
\setlength{\tabcolsep}{2.1mm}{
\begin{tabular}{l|ccc}
\toprule
Method & MME$^{P}$ & MMBench & SEED$^{I}$ \\ \midrule
Basline & 1516.9 & 65.3 & 66.8 \\
+BLIP-558K & 1521.6 & 66.2 & 66.9 \\
+ShareGPT4V-PT-558K & \textbf{1539.8} & \textbf{68.3} & \textbf{68.9} \\ \bottomrule
\end{tabular}
}
\vspace{-3mm}
\caption{
\textbf{Ablation on the pre-training caption quality.} Based on the baseline, the second and third rows share the same end-to-end training strategy and images, but different captions from the BLIP captioner or our ShareGPT4V-PT dataset.
}\label{tab:pt_cap_source}
\vspace{-4mm}
\end{table}

\noindent{\textbf{Pre-training Caption Quality.}}
Then we study how the caption quality influences the pre-training performance. For a fair comparison, we pre-train the model with the same setting and images, but the captions are generated by different models.
In detail, we use the 558K LAION-CC-SUB image-text pairs captioned by the BLIP as the baseline and replace the text with the high-quality one in our ShareGPT4V-PT. 

As results shown in Table \ref{tab:pt_cap_source}, comparing with the baseline, the joint training strategy with the BLIP-558K data gets better results on all the benchmarks, while the gain is quite minor that only $4.7$ in MME Perception and $0.1$ on SEED Bench. When we replace the captions with our ShareGPT4V-PT-558K, the model gets significant gains. In detail, it gets $1549.8$, $68.3$, $68.9$ on the three benchmarks, surpassing the BLIP-558K case with $18.2$, $1.9$ and $2.0$ respectively. This proves the essential of high-quality captions for effective pre-training and modality alignment.

\begin{figure}[t]
\centering
\includegraphics[width=1\linewidth]{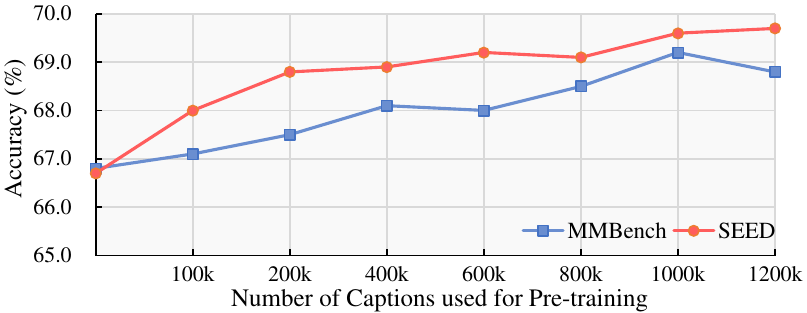}
\vspace{-7mm}
\caption{
The pre-training data scaling performance on MMBench and SEED Bench. The model shows consistent gain with more pre-training data.
}
\label{fig:num_cap}
\vspace{-4mm}
\end{figure}

\noindent{\textbf{Number of Captions in Pre-training.}} 
In Figure \ref{fig:num_cap}, we present our investigation into the required quantity of high-quality captions for the pre-training stage. Here we randomly sample the data from the ShareGPT4V-PT and train the model with the subset, which varies from 100K to 1200K.
The results show that with only 100K high-quality data, the model has a significant improvement on both benchmarks, this further proves the effectiveness of the high-quality data. 
Meanwhile, with the scaling of training data, the model performance tends to be saturated after more than 1000K data being used for pre-training. This may indicate that with high-quality captions, the modal alignment could be quite efficient and realized with a relatively lightweight data scale.

\noindent{\textbf{Number of Learnable ViT Blocks in Pre-training.}} As detailed in Table \ref{tab:ablation_vit_layer}, we extensively investigated the optimal approach for fine-tuning the vision encoder during the pretraining phase. Compared to freezing the vision encoder during the pretraining phase, we found that unlocking the latter half of its transformer blocks significantly enhances performance. Specifically, such an approach led to a 52.2 gain on the MME perception benchmark, and substantial accuracy improvements of 2.2$\%$ and 1.6$\%$ on the MMBench and SEED benchmarks, respectively. This suggests that for high-quality captions, unlocking the vision encoder facilitates more effective modality alignment. 

\begin{table}[h!]
\centering
\small
\vspace{-2mm}
\setlength{\tabcolsep}{1.4mm}{
\begin{tabular}{cc|lll}
\toprule
Tune from Block & Memory Usage & MME$^{P}$ & MMB & SEED$^{I}$ \\ \midrule
24 &49.6 GB & 1515.2 & 66.6 & 68.1 \\
18 &53.2 GB & 1556.0 & 67.2 & 69.3 \\
12 &56.7 GB &\textbf{1567.4} & \textbf{68.8} & \textbf{69.7} \\
6 &60.0 GB & 1529.5  & 67.7 & 69.6 \\
0 &63.6 GB & 1545.7 & 68.5 & 69.2 \\ \bottomrule
\end{tabular}
}
\vspace{-3mm}
\caption{Ablation study about the number of learnable blocks in the vision encoder.}\label{tab:ablation_vit_layer}
\vspace{-5mm}
\end{table}

\section{Conclusion}
In this study, we introduce ShareGPT4V, a groundbreaking large-scale image-text dataset with 1.2 million detailed and informative captions that surpass existing datasets in terms of richness and diversity, covering world knowledge, object attributes, spatial relationships, and aesthetic assessments. ShareGPT4V comprises 100K high-quality captions from GPT4-Vision for Supervised Fine-Tuning (SFT), expanded to 1.2 million for pre-training through a general caption model. We validated ShareGPT4V's effectiveness through SFT results on recent LMMs and further demonstrated its capabilities with the superior performance of our ShareGPT4V-7B model, which incorporates the dataset in both pre-training and SFT stages. We are committed to making ShareGPT4V fully accessible to the public, with the aspiration that it becomes a foundational resource in advancing the field of LMMs.

\appendix

\section{Data Sources}
\label{sec:data_sources}

\textbf{Data Source Composition for ShareGPT4V.} To maximize the comprehensiveness of our captions, we compiled a total of 100K images from diverse sources. This includes 50K images from COCO \cite{lin2014microsoft}, 30K images from 'LCS'  (which abbreviates LAION \cite{schuhmann2021laion}, CC-3M \cite{sharma2018conceptual}, and SBU \cite{ordonez2011im2text}), 20K images from SAM \cite{kirillov2023segment}, 500 images from TextCaps \cite{sidorov2020textcaps}, 500 images from WikiArt \cite{saleh2015large}, and 1K images from web-crawled data (split evenly between images of landmarks and images of celebrities).

\vspace{2mm}
\noindent{\textbf{Data Source Composition for ShareGPT4V-PT.}}  We utilized our pre-trained Share-Captioner to generate the pre-training dataset. This dataset is comprised of a subset of 1.2M images selected from existing public datasets. These include 118K images from COCO \cite{lin2014microsoft},570K images from SAM \cite{kirillov2023segment}, and 558K images from LLaVA-1.5 pre-training data \cite{liu2023improved}.

\section{Caption Analysis}
\begin{figure}
\centering
\includegraphics[width=0.9\linewidth]{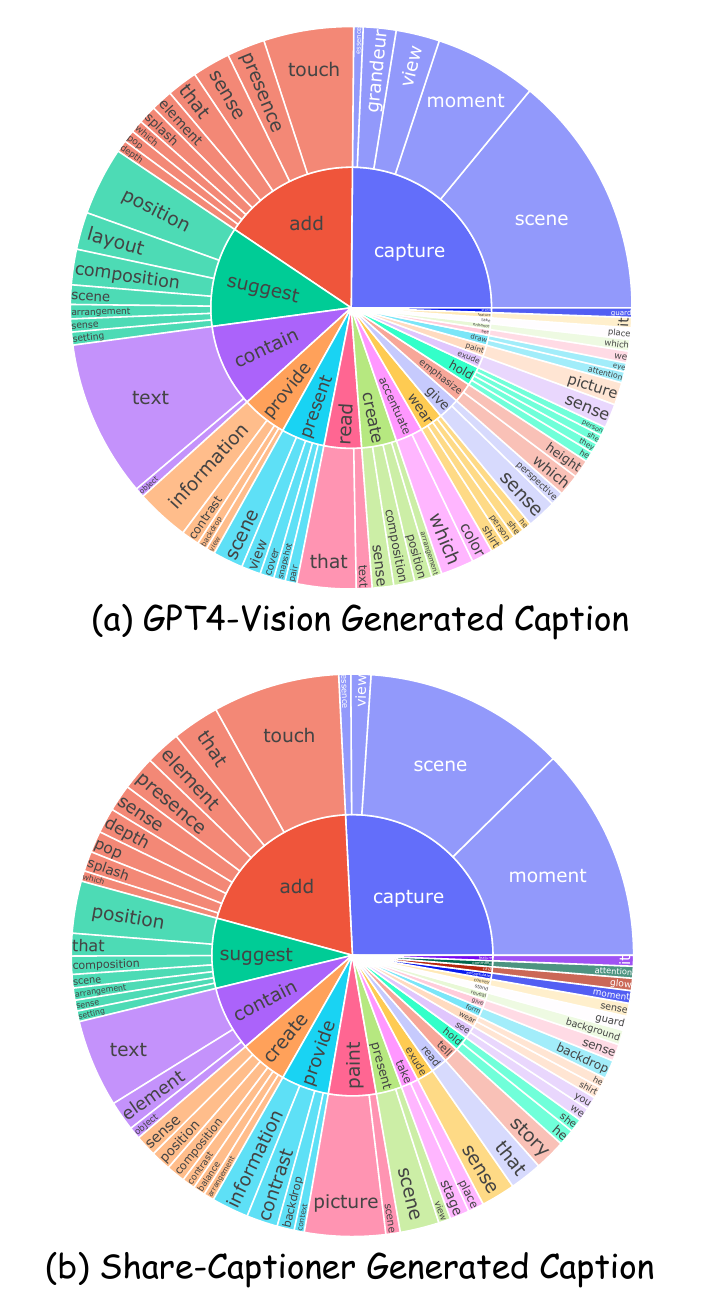}
\vspace{-2mm}
\caption{
 \textbf{Analysis of captions generated by GPT4-Vision and Share-Captioner}. Visualization of the root noun-verb pairs (occurring over 1\%) of the captions.
}
\vspace{2mm}
\label{fig:cap_diversity}
\end{figure}

Figure \ref{fig:cap_diversity} provides a visualization of the root noun-verb pairs for the captions generated by both GPT4-Vision and Share-Captioner. It’s clear to see that the diversity and linguistic expression of the captions produced by Share-Captioner are comparable to those of GPT4-Vision.

We analyzed the lexical composition of the captions produced by GPT4-Vision and Share-Captioner, and the results are presented in Table \ref{table:lexical}. The analysis reveals that the captions generated by our Share-Captioner contain a comparable amount of information to those generated by GPT4-Vision.

\begin{table}[h]
\centering
\resizebox{1\linewidth}{!}{
\setlength{\tabcolsep}{1mm}{
\begin{tabular}{l|cccccc}
\toprule
Lexical &n. &adj. &adv. &v. &num. &prep.\\ \midrule
GPT4-Vision &27.3$\%$ &9.5$\%$ &2.0$\%$ &12.3$\%$ &0.5$\%$ &11.4$\%$ \\ \midrule
Share-Captioner &27.4$\%$ &8.8$\%$ &1.5$\%$ &12.5$\%$ &0.4$\%$ &11.5$\%$ \\ \bottomrule
\end{tabular}
}}
\vspace{-2mm}
\caption{\textbf{Comparison of lexical composition of the captions} generated by GPT4-Vision and Share-Captioner.} \label{table:lexical}
\end{table}

\section{Prompts}

\begin{figure*}[!t]
\centering
\includegraphics[width=1\textwidth]{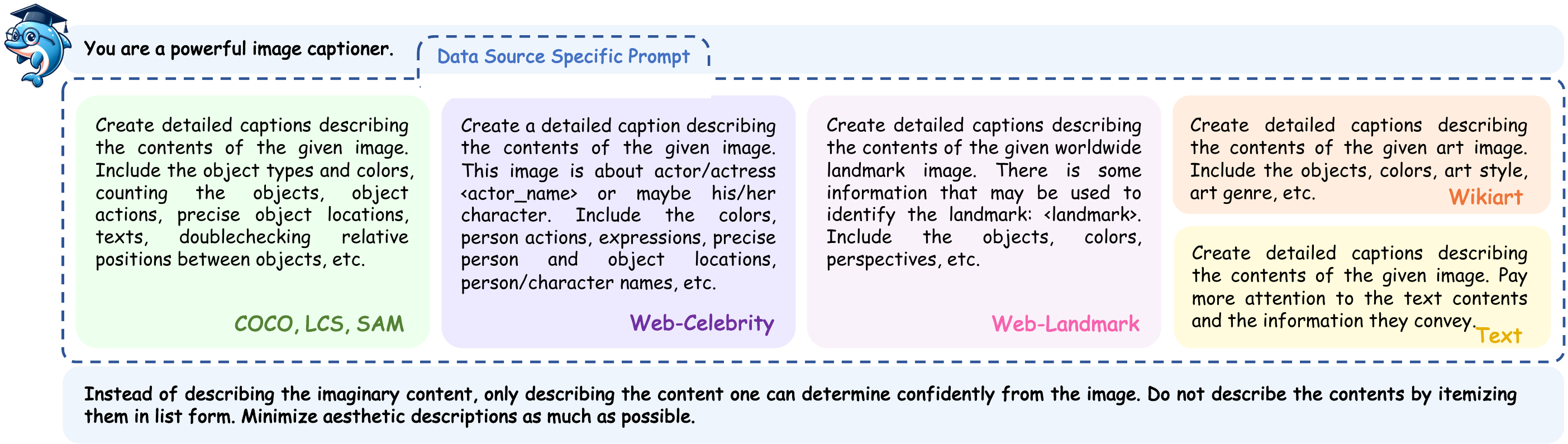}
\caption{
\textbf{Prompts for instructing GPT4-Vision to generate detailed descriptions}. The Prompts are designed with base prompts at the beginning and end, with a data-specific prompt placed in between. 
}
\vspace{-20mm}
\label{fig:data_prompt}
\end{figure*}

Given the diversity of our image sources, we expect a highly content-related description for each image. As shown in Figure \ref{fig:data_prompt}, we designed a base prompt for a general description and added a specialized prompt for each data source. 

\section{Examples}

\begin{figure*}
\centering
\includegraphics[width=1\textwidth]{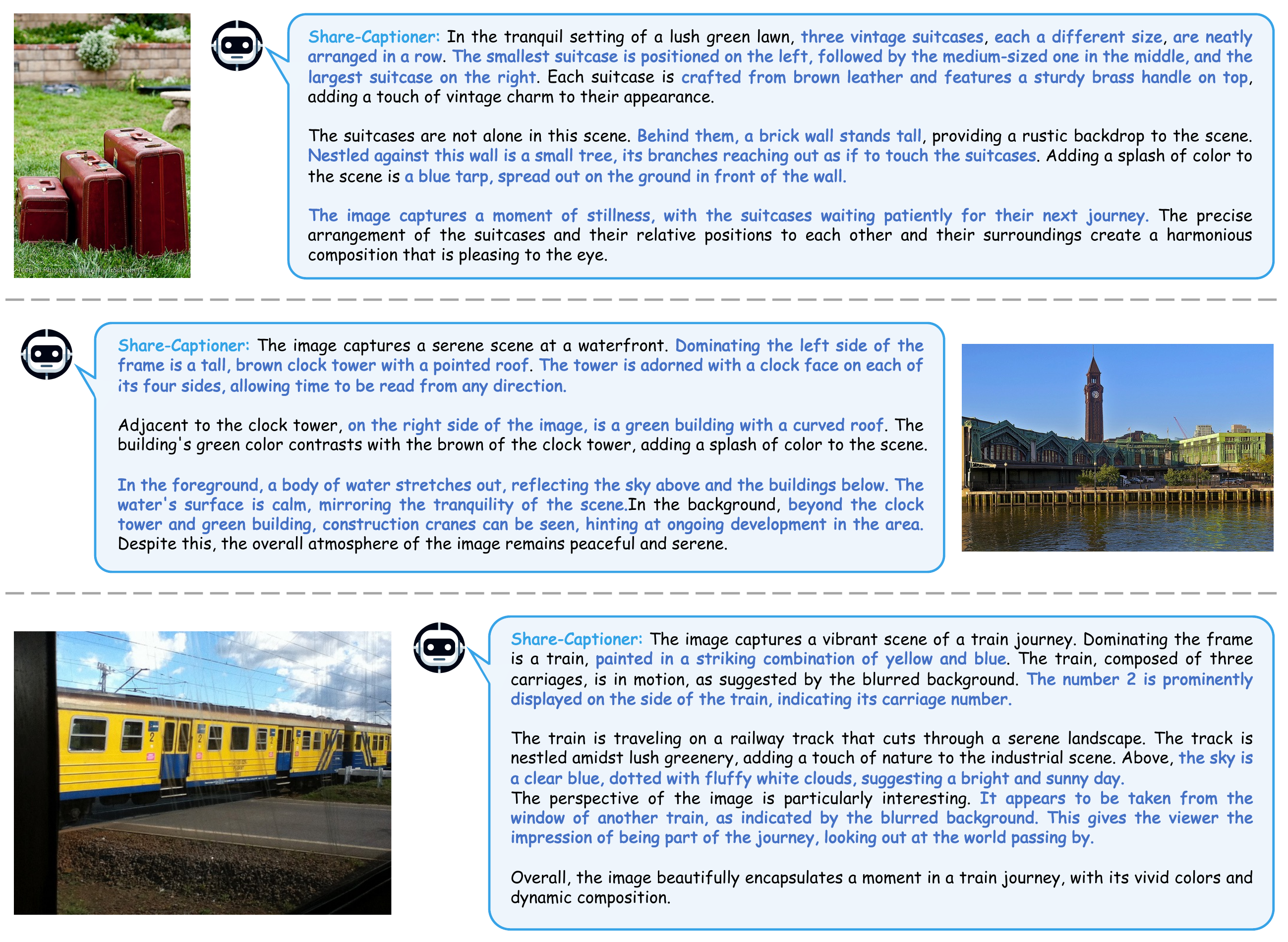}
\caption{
\textbf{Examples of captions generated by Share-Captioner}. Detailed and accurate parts within the captions are emphasized in \textcolor{shareblue}{blue}.
}
\vspace{-3mm}
\label{examples_captioner}
\end{figure*}

\begin{figure*}
\centering
\includegraphics[width=1\textwidth]{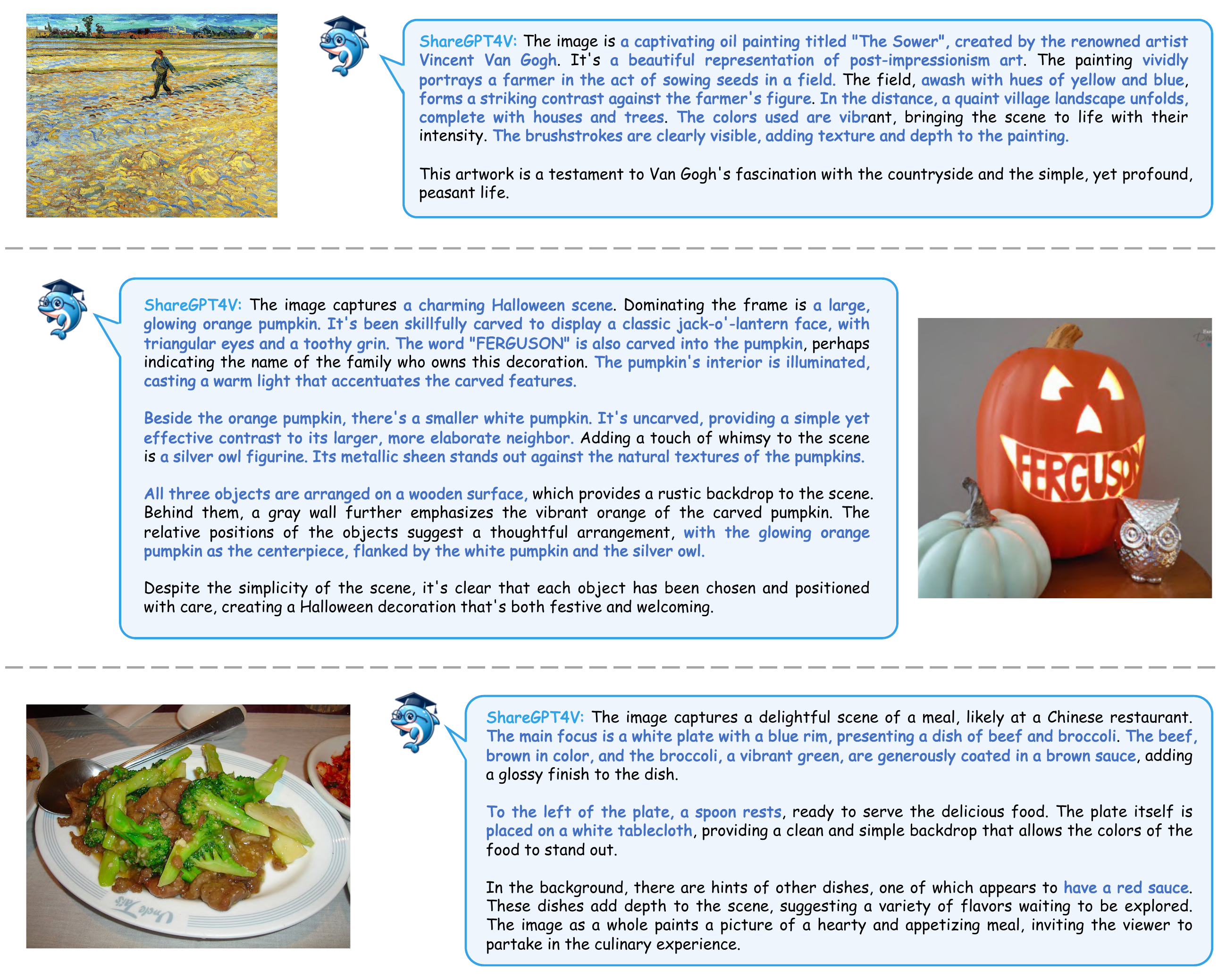}
\caption{
\textbf{Examples of captions generated by ShareGPT4V}. Detailed and accurate parts within the captions are emphasized in \textcolor{shareblue}{blue}. 
}
\vspace{-3mm}
\label{examples_4v}
\end{figure*}

\begin{figure*}
\centering
\includegraphics[width=1\textwidth]{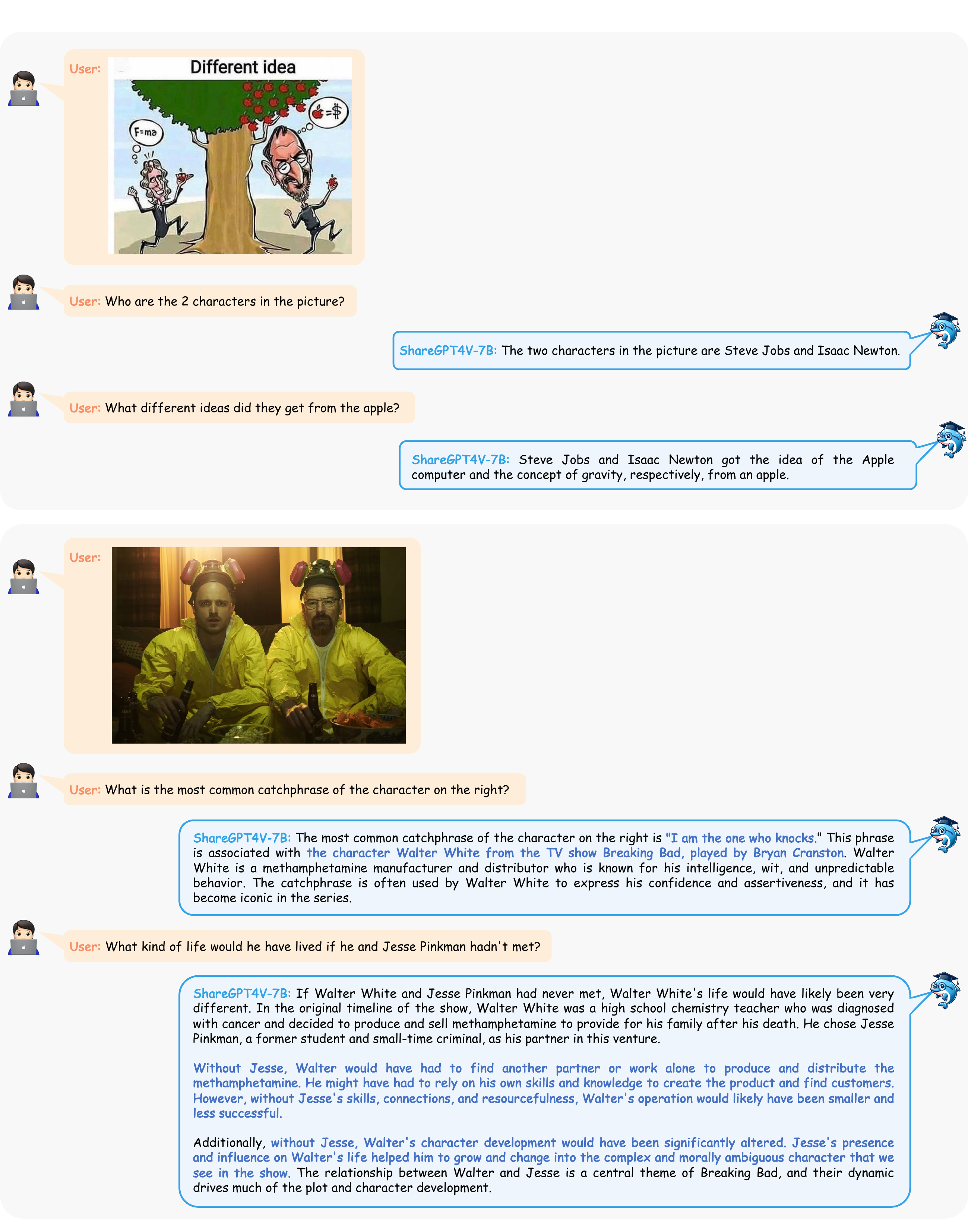}
\caption{
\textbf{Example of Multi-Round Dialog of ShareGPT4V-7B}. Detailed and accurate parts within the captions are emphasized in \textcolor{shareblue}{blue}. 
}
\vspace{-3mm}
\label{examples_multi_2}
\end{figure*}
\clearpage
\clearpage
{
    \small
    \bibliographystyle{ieeenat_fullname}
    \bibliography{main}

\begin{thebibliography}{62}
\providecommand{\natexlab}[1]{#1}
\providecommand{\url}[1]{\texttt{#1}}
\expandafter\ifx\csname urlstyle\endcsname\relax
  \providecommand{\doi}[1]{doi: #1}\else
  \providecommand{\doi}{doi: \begingroup \urlstyle{rm}\Url}\fi

\bibitem[sha(2023)]{sharegpt}
Sharegpt.
\newblock \url{https://sharegpt.com/}, 2023.

\bibitem[Bai et~al.(2023{\natexlab{a}})Bai, Bai, Chu, Cui, Dang, Deng, Fan, Ge, Han, Huang, et~al.]{bai2023qwen}
Jinze Bai, Shuai Bai, Yunfei Chu, Zeyu Cui, Kai Dang, Xiaodong Deng, Yang Fan, Wenbin Ge, Yu Han, Fei Huang, et~al.
\newblock Qwen technical report.
\newblock \emph{arXiv preprint arXiv:2309.16609}, 2023{\natexlab{a}}.

\bibitem[Bai et~al.(2023{\natexlab{b}})Bai, Bai, Yang, Wang, Tan, Wang, Lin, Zhou, and Zhou]{bai2023qwenvl}
Jinze Bai, Shuai Bai, Shusheng Yang, Shijie Wang, Sinan Tan, Peng Wang, Junyang Lin, Chang Zhou, and Jingren Zhou.
\newblock Qwen-vl: A frontier large vision-language model with versatile abilities.
\newblock \emph{arXiv preprint arXiv:2308.12966}, 2023{\natexlab{b}}.

\bibitem[Brown et~al.(2020)Brown, Mann, Ryder, Subbiah, Kaplan, Dhariwal, Neelakantan, Shyam, Sastry, Askell, et~al.]{brown2020language}
Tom Brown, Benjamin Mann, Nick Ryder, Melanie Subbiah, Jared~D Kaplan, Prafulla Dhariwal, Arvind Neelakantan, Pranav Shyam, Girish Sastry, Amanda Askell, et~al.
\newblock Language models are few-shot learners.
\newblock \emph{Advances in neural information processing systems}, 33:\penalty0 1877--1901, 2020.

\bibitem[Chen et~al.(2023{\natexlab{a}})Chen, Li, Zhang, Xiong, and Elhoseiny]{chen2023minigpt}
Jun Chen, Deyao Zhu1 Xiaoqian Shen1~Xiang Li, Zechun Liu2~Pengchuan Zhang, Raghuraman Krishnamoorthi2 Vikas Chandra2~Yunyang Xiong, and Mohamed Elhoseiny.
\newblock Minigpt-v2: Large language model as a unified interface for vision-language multi-task learning.
\newblock \emph{arXiv preprint arXiv:2310.09478}, 2023{\natexlab{a}}.

\bibitem[Chen et~al.(2023{\natexlab{b}})Chen, Zhang, Zeng, Zhang, Zhu, and Zhao]{chen2023shikra}
Keqin Chen, Zhao Zhang, Weili Zeng, Richong Zhang, Feng Zhu, and Rui Zhao.
\newblock Shikra: Unleashing multimodal llm's referential dialogue magic.
\newblock \emph{arXiv preprint arXiv:2306.15195}, 2023{\natexlab{b}}.

\bibitem[Chen et~al.(2015)Chen, Fang, Lin, Vedantam, Gupta, Doll{\'a}r, and Zitnick]{chen2015microsoft}
Xinlei Chen, Hao Fang, Tsung-Yi Lin, Ramakrishna Vedantam, Saurabh Gupta, Piotr Doll{\'a}r, and C~Lawrence Zitnick.
\newblock Microsoft coco captions: Data collection and evaluation server.
\newblock \emph{arXiv preprint arXiv:1504.00325}, 2015.

\bibitem[Chiang et~al.(2023)Chiang, Li, Lin, Sheng, Wu, Zhang, Zheng, Zhuang, Zhuang, Gonzalez, et~al.]{chiang2023vicuna}
Wei-Lin Chiang, Zhuohan Li, Zi Lin, Ying Sheng, Zhanghao Wu, Hao Zhang, Lianmin Zheng, Siyuan Zhuang, Yonghao Zhuang, Joseph~E Gonzalez, et~al.
\newblock Vicuna: An open-source chatbot impressing gpt-4 with 90\%* chatgpt quality.
\newblock \emph{See https://vicuna. lmsys. org (accessed 14 April 2023)}, 2023.

\bibitem[Chowdhery et~al.(2022)Chowdhery, Narang, Devlin, Bosma, Mishra, Roberts, Barham, Chung, Sutton, Gehrmann, et~al.]{chowdhery2022palm}
Aakanksha Chowdhery, Sharan Narang, Jacob Devlin, Maarten Bosma, Gaurav Mishra, Adam Roberts, Paul Barham, Hyung~Won Chung, Charles Sutton, Sebastian Gehrmann, et~al.
\newblock Palm: Scaling language modeling with pathways.
\newblock \emph{arXiv preprint arXiv:2204.02311}, 2022.

\bibitem[Dai et~al.(2023)Dai, Li, Li, Tiong, Zhao, Wang, Li, Fung, and Hoi]{instructblip}
Wenliang Dai, Junnan Li, Dongxu Li, Anthony Meng~Huat Tiong, Junqi Zhao, Weisheng Wang, Boyang Li, Pascale Fung, and Steven Hoi.
\newblock Instructblip: Towards general-purpose vision-language models with instruction tuning, 2023.

\bibitem[Devlin et~al.(2018)Devlin, Chang, Lee, and Toutanova]{devlin2018bert}
Jacob Devlin, Ming-Wei Chang, Kenton Lee, and Kristina Toutanova.
\newblock Bert: Pre-training of deep bidirectional transformers for language understanding.
\newblock \emph{arXiv preprint arXiv:1810.04805}, 2018.

\bibitem[Du et~al.(2021)Du, Qian, Liu, Ding, Qiu, Yang, and Tang]{du2021glm}
Zhengxiao Du, Yujie Qian, Xiao Liu, Ming Ding, Jiezhong Qiu, Zhilin Yang, and Jie Tang.
\newblock Glm: General language model pretraining with autoregressive blank infilling.
\newblock \emph{arXiv preprint arXiv:2103.10360}, 2021.

\bibitem[Fan et~al.(2023)Fan, Krishnan, Isola, Katabi, and Tian]{fan2023improving}
Lijie Fan, Dilip Krishnan, Phillip Isola, Dina Katabi, and Yonglong Tian.
\newblock Improving clip training with language rewrites.
\newblock \emph{arXiv preprint arXiv:2305.20088}, 2023.

\bibitem[Fang et~al.(2023)Fang, Wang, Xie, Sun, Wu, Wang, Huang, Wang, and Cao]{fang2023eva}
Yuxin Fang, Wen Wang, Binhui Xie, Quan Sun, Ledell Wu, Xinggang Wang, Tiejun Huang, Xinlong Wang, and Yue Cao.
\newblock Eva: Exploring the limits of masked visual representation learning at scale.
\newblock In \emph{Proceedings of the IEEE/CVF Conference on Computer Vision and Pattern Recognition}, pages 19358--19369, 2023.

\bibitem[Fu et~al.(2023)Fu, Chen, Shen, Qin, Zhang, Lin, Qiu, Lin, Yang, Zheng, Li, Sun, and Ji]{fu2023mme}
Chaoyou Fu, Peixian Chen, Yunhang Shen, Yulei Qin, Mengdan Zhang, Xu Lin, Zhenyu Qiu, Wei Lin, Jinrui Yang, Xiawu Zheng, Ke Li, Xing Sun, and Rongrong Ji.
\newblock Mme: A comprehensive evaluation benchmark for multimodal large language models.
\newblock \emph{arXiv preprint arXiv:2306.13394}, 2023.

\bibitem[Gadre et~al.(2023)Gadre, Ilharco, Fang, Hayase, Smyrnis, Nguyen, Marten, Wortsman, Ghosh, Zhang, et~al.]{gadre2023datacomp}
Samir~Yitzhak Gadre, Gabriel Ilharco, Alex Fang, Jonathan Hayase, Georgios Smyrnis, Thao Nguyen, Ryan Marten, Mitchell Wortsman, Dhruba Ghosh, Jieyu Zhang, et~al.
\newblock Datacomp: In search of the next generation of multimodal datasets.
\newblock \emph{arXiv preprint arXiv:2304.14108}, 2023.

\bibitem[Goyal et~al.(2017)Goyal, Khot, Summers-Stay, Batra, and Parikh]{goyal2017making}
Yash Goyal, Tejas Khot, Douglas Summers-Stay, Dhruv Batra, and Devi Parikh.
\newblock Making the v in vqa matter: Elevating the role of image understanding in visual question answering.
\newblock In \emph{Proceedings of the IEEE conference on computer vision and pattern recognition}, pages 6904--6913, 2017.

\bibitem[Gurari et~al.(2018)Gurari, Li, Stangl, Guo, Lin, Grauman, Luo, and Bigham]{gurari2018vizwiz}
Danna Gurari, Qing Li, Abigale~J Stangl, Anhong Guo, Chi Lin, Kristen Grauman, Jiebo Luo, and Jeffrey~P Bigham.
\newblock Vizwiz grand challenge: Answering visual questions from blind people.
\newblock In \emph{Proceedings of the IEEE conference on computer vision and pattern recognition}, pages 3608--3617, 2018.

\bibitem[Jia et~al.(2021)Jia, Yang, Xia, Chen, Parekh, Pham, Le, Sung, Li, and Duerig]{jia2021scaling}
Chao Jia, Yinfei Yang, Ye Xia, Yi-Ting Chen, Zarana Parekh, Hieu Pham, Quoc Le, Yun-Hsuan Sung, Zhen Li, and Tom Duerig.
\newblock Scaling up visual and vision-language representation learning with noisy text supervision.
\newblock In \emph{International conference on machine learning}, pages 4904--4916. PMLR, 2021.

\bibitem[Kazemzadeh et~al.(2014)Kazemzadeh, Ordonez, Matten, and Berg]{kazemzadeh2014referitgame}
Sahar Kazemzadeh, Vicente Ordonez, Mark Matten, and Tamara Berg.
\newblock Referitgame: Referring to objects in photographs of natural scenes.
\newblock In \emph{Proceedings of the 2014 conference on empirical methods in natural language processing (EMNLP)}, pages 787--798, 2014.

\bibitem[Kirillov et~al.(2023)Kirillov, Mintun, Ravi, Mao, Rolland, Gustafson, Xiao, Whitehead, Berg, Lo, et~al.]{kirillov2023segment}
Alexander Kirillov, Eric Mintun, Nikhila Ravi, Hanzi Mao, Chloe Rolland, Laura Gustafson, Tete Xiao, Spencer Whitehead, Alexander~C Berg, Wan-Yen Lo, et~al.
\newblock Segment anything.
\newblock \emph{arXiv preprint arXiv:2304.02643}, 2023.

\bibitem[Krishna et~al.(2017)Krishna, Zhu, Groth, Johnson, Hata, Kravitz, Chen, Kalantidis, Li, Shamma, et~al.]{krishna2017visual}
Ranjay Krishna, Yuke Zhu, Oliver Groth, Justin Johnson, Kenji Hata, Joshua Kravitz, Stephanie Chen, Yannis Kalantidis, Li-Jia Li, David~A Shamma, et~al.
\newblock Visual genome: Connecting language and vision using crowdsourced dense image annotations.
\newblock \emph{International journal of computer vision}, 123:\penalty0 32--73, 2017.

\bibitem[Lai et~al.(2023)Lai, Zhang, Wu, Bai, Timofeev, Du, Gan, Shan, Chuah, Yang, et~al.]{lai2023scarcity}
Zhengfeng Lai, Haotian Zhang, Wentao Wu, Haoping Bai, Aleksei Timofeev, Xianzhi Du, Zhe Gan, Jiulong Shan, Chen-Nee Chuah, Yinfei Yang, et~al.
\newblock From scarcity to efficiency: Improving clip training via visual-enriched captions.
\newblock \emph{arXiv preprint arXiv:2310.07699}, 2023.

\bibitem[Li et~al.(2023{\natexlab{a}})Li, Wang, Wang, Ge, Ge, and Shan]{li2023seed}
Bohao Li, Rui Wang, Guangzhi Wang, Yuying Ge, Yixiao Ge, and Ying Shan.
\newblock Seed-bench: Benchmarking multimodal llms with generative comprehension.
\newblock \emph{arXiv preprint arXiv:2307.16125}, 2023{\natexlab{a}}.

\bibitem[Li et~al.(2023{\natexlab{b}})Li, Zhang, Chen, Wang, Yang, and Liu]{li2023otter}
Bo Li, Yuanhan Zhang, Liangyu Chen, Jinghao Wang, Jingkang Yang, and Ziwei Liu.
\newblock Otter: A multi-modal model with in-context instruction tuning.
\newblock \emph{arXiv preprint arXiv:2305.03726}, 2023{\natexlab{b}}.

\bibitem[Li et~al.(2022{\natexlab{a}})Li, Li, Xiong, and Hoi]{li2022blip}
Junnan Li, Dongxu Li, Caiming Xiong, and Steven Hoi.
\newblock Blip: Bootstrapping language-image pre-training for unified vision-language understanding and generation.
\newblock In \emph{International Conference on Machine Learning}, pages 12888--12900. PMLR, 2022{\natexlab{a}}.

\bibitem[Li et~al.(2023{\natexlab{c}})Li, Li, Savarese, and Hoi]{li2023blip}
Junnan Li, Dongxu Li, Silvio Savarese, and Steven Hoi.
\newblock Blip-2: Bootstrapping language-image pre-training with frozen image encoders and large language models.
\newblock \emph{arXiv preprint arXiv:2301.12597}, 2023{\natexlab{c}}.

\bibitem[Li et~al.(2022{\natexlab{b}})Li, Zhang, Zhang, Yang, Li, Zhong, Wang, Yuan, Zhang, Hwang, et~al.]{li2022grounded}
Liunian~Harold Li, Pengchuan Zhang, Haotian Zhang, Jianwei Yang, Chunyuan Li, Yiwu Zhong, Lijuan Wang, Lu Yuan, Lei Zhang, Jenq-Neng Hwang, et~al.
\newblock Grounded language-image pre-training.
\newblock In \emph{Proceedings of the IEEE/CVF Conference on Computer Vision and Pattern Recognition}, pages 10965--10975, 2022{\natexlab{b}}.

\bibitem[Lin et~al.(2014)Lin, Maire, Belongie, Hays, Perona, Ramanan, Doll{\'a}r, and Zitnick]{lin2014microsoft}
Tsung-Yi Lin, Michael Maire, Serge Belongie, James Hays, Pietro Perona, Deva Ramanan, Piotr Doll{\'a}r, and C~Lawrence Zitnick.
\newblock Microsoft coco: Common objects in context.
\newblock In \emph{Computer Vision--ECCV 2014: 13th European Conference, Zurich, Switzerland, September 6-12, 2014, Proceedings, Part V 13}, pages 740--755. Springer, 2014.

\bibitem[Liu et~al.(2023{\natexlab{a}})Liu, Li, Li, and Lee]{liu2023improved}
Haotian Liu, Chunyuan Li, Yuheng Li, and Yong~Jae Lee.
\newblock Improved baselines with visual instruction tuning.
\newblock \emph{arXiv preprint arXiv:2310.03744}, 2023{\natexlab{a}}.

\bibitem[Liu et~al.(2023{\natexlab{b}})Liu, Li, Wu, and Lee]{liu2023visual}
Haotian Liu, Chunyuan Li, Qingyang Wu, and Yong~Jae Lee.
\newblock Visual instruction tuning.
\newblock \emph{arXiv preprint arXiv:2304.08485}, 2023{\natexlab{b}}.

\bibitem[Liu et~al.(2023{\natexlab{c}})Liu, Zeng, Ren, Li, Zhang, Yang, Li, Yang, Su, Zhu, et~al.]{liu2023grounding}
Shilong Liu, Zhaoyang Zeng, Tianhe Ren, Feng Li, Hao Zhang, Jie Yang, Chunyuan Li, Jianwei Yang, Hang Su, Jun Zhu, et~al.
\newblock Grounding dino: Marrying dino with grounded pre-training for open-set object detection.
\newblock \emph{arXiv preprint arXiv:2303.05499}, 2023{\natexlab{c}}.

\bibitem[Liu et~al.(2023{\natexlab{d}})Liu, Duan, Zhang, Li, Zhang, Zhao, Yuan, Wang, He, Liu, et~al.]{liu2023mmbench}
Yuan Liu, Haodong Duan, Yuanhan Zhang, Bo Li, Songyang Zhang, Wangbo Zhao, Yike Yuan, Jiaqi Wang, Conghui He, Ziwei Liu, et~al.
\newblock Mmbench: Is your multi-modal model an all-around player?
\newblock \emph{arXiv preprint arXiv:2307.06281}, 2023{\natexlab{d}}.

\bibitem[Lu et~al.(2022)Lu, Mishra, Xia, Qiu, Chang, Zhu, Tafjord, Clark, and Kalyan]{lu2022learn}
Pan Lu, Swaroop Mishra, Tanglin Xia, Liang Qiu, Kai-Wei Chang, Song-Chun Zhu, Oyvind Tafjord, Peter Clark, and Ashwin Kalyan.
\newblock Learn to explain: Multimodal reasoning via thought chains for science question answering.
\newblock \emph{Advances in Neural Information Processing Systems}, 35:\penalty0 2507--2521, 2022.

\bibitem[Luo et~al.(2023)Luo, Zhou, Ren, Chen, Sun, and Ji]{luo2023cheap}
Gen Luo, Yiyi Zhou, Tianhe Ren, Shengxin Chen, Xiaoshuai Sun, and Rongrong Ji.
\newblock Cheap and quick: Efficient vision-language instruction tuning for large language models.
\newblock \emph{arXiv preprint arXiv:2305.15023}, 2023.

\bibitem[Marino et~al.(2019)Marino, Rastegari, Farhadi, and Mottaghi]{marino2019ok}
Kenneth Marino, Mohammad Rastegari, Ali Farhadi, and Roozbeh Mottaghi.
\newblock Ok-vqa: A visual question answering benchmark requiring external knowledge.
\newblock In \emph{Proceedings of the IEEE/cvf conference on computer vision and pattern recognition}, pages 3195--3204, 2019.

\bibitem[Mishra et~al.(2019)Mishra, Shekhar, Singh, and Chakraborty]{mishra2019ocr}
Anand Mishra, Shashank Shekhar, Ajeet~Kumar Singh, and Anirban Chakraborty.
\newblock Ocr-vqa: Visual question answering by reading text in images.
\newblock In \emph{2019 international conference on document analysis and recognition (ICDAR)}, pages 947--952. IEEE, 2019.

\bibitem[Nguyen et~al.(2023)Nguyen, Gadre, Ilharco, Oh, and Schmidt]{nguyen2023improving}
Thao Nguyen, Samir~Yitzhak Gadre, Gabriel Ilharco, Sewoong Oh, and Ludwig Schmidt.
\newblock Improving multimodal datasets with image captioning.
\newblock \emph{arXiv preprint arXiv:2307.10350}, 2023.

\bibitem[OpenAI(2023{\natexlab{a}})]{chatgpt}
OpenAI.
\newblock Chatgpt.
\newblock \url{https://chat.openai.com/}, 2023{\natexlab{a}}.

\bibitem[OpenAI(2023{\natexlab{b}})]{gpt4v}
OpenAI.
\newblock Gpt-4v(ision) system card.
\newblock 2023{\natexlab{b}}.

\bibitem[Ordonez et~al.(2011)Ordonez, Kulkarni, and Berg]{ordonez2011im2text}
Vicente Ordonez, Girish Kulkarni, and Tamara Berg.
\newblock Im2text: Describing images using 1 million captioned photographs.
\newblock \emph{Advances in neural information processing systems}, 24, 2011.

\bibitem[Ouyang et~al.(2022)Ouyang, Wu, Jiang, Almeida, Wainwright, Mishkin, Zhang, Agarwal, Slama, Ray, et~al.]{ouyang2022training}
Long Ouyang, Jeffrey Wu, Xu Jiang, Diogo Almeida, Carroll Wainwright, Pamela Mishkin, Chong Zhang, Sandhini Agarwal, Katarina Slama, Alex Ray, et~al.
\newblock Training language models to follow instructions with human feedback.
\newblock \emph{Advances in Neural Information Processing Systems}, 35:\penalty0 27730--27744, 2022.

\bibitem[Peng et~al.(2023)Peng, Wang, Dong, Hao, Huang, Ma, and Wei]{peng2023kosmos}
Zhiliang Peng, Wenhui Wang, Li Dong, Yaru Hao, Shaohan Huang, Shuming Ma, and Furu Wei.
\newblock Kosmos-2: Grounding multimodal large language models to the world.
\newblock \emph{arXiv preprint arXiv:2306.14824}, 2023.

\bibitem[Radford et~al.(2018)Radford, Narasimhan, Salimans, Sutskever, et~al.]{radford2018improving}
Alec Radford, Karthik Narasimhan, Tim Salimans, Ilya Sutskever, et~al.
\newblock Improving language understanding by generative pre-training.
\newblock 2018.

\bibitem[Radford et~al.(2021)Radford, Kim, Hallacy, Ramesh, Goh, Agarwal, Sastry, Askell, Mishkin, Clark, et~al.]{radford2021learning}
Alec Radford, Jong~Wook Kim, Chris Hallacy, Aditya Ramesh, Gabriel Goh, Sandhini Agarwal, Girish Sastry, Amanda Askell, Pamela Mishkin, Jack Clark, et~al.
\newblock Learning transferable visual models from natural language supervision.
\newblock In \emph{International conference on machine learning}, pages 8748--8763. PMLR, 2021.

\bibitem[Raffel et~al.(2020)Raffel, Shazeer, Roberts, Lee, Narang, Matena, Zhou, Li, and Liu]{raffel2020exploring}
Colin Raffel, Noam Shazeer, Adam Roberts, Katherine Lee, Sharan Narang, Michael Matena, Yanqi Zhou, Wei Li, and Peter~J Liu.
\newblock Exploring the limits of transfer learning with a unified text-to-text transformer.
\newblock \emph{The Journal of Machine Learning Research}, 21\penalty0 (1):\penalty0 5485--5551, 2020.

\bibitem[Saleh and Elgammal(2015)]{saleh2015large}
Babak Saleh and Ahmed Elgammal.
\newblock Large-scale classification of fine-art paintings: Learning the right metric on the right feature.
\newblock \emph{arXiv preprint arXiv:1505.00855}, 2015.

\bibitem[Schuhmann et~al.(2021)Schuhmann, Vencu, Beaumont, Kaczmarczyk, Mullis, Katta, Coombes, Jitsev, and Komatsuzaki]{schuhmann2021laion}
Christoph Schuhmann, Richard Vencu, Romain Beaumont, Robert Kaczmarczyk, Clayton Mullis, Aarush Katta, Theo Coombes, Jenia Jitsev, and Aran Komatsuzaki.
\newblock Laion-400m: Open dataset of clip-filtered 400 million image-text pairs.
\newblock \emph{arXiv preprint arXiv:2111.02114}, 2021.

\bibitem[Schwenk et~al.(2022)Schwenk, Khandelwal, Clark, Marino, and Mottaghi]{schwenk2022okvqa}
Dustin Schwenk, Apoorv Khandelwal, Christopher Clark, Kenneth Marino, and Roozbeh Mottaghi.
\newblock A-okvqa: A benchmark for visual question answering using world knowledge.
\newblock In \emph{European Conference on Computer Vision}, pages 146--162. Springer, 2022.

\bibitem[Sharma et~al.(2018)Sharma, Ding, Goodman, and Soricut]{sharma2018conceptual}
Piyush Sharma, Nan Ding, Sebastian Goodman, and Radu Soricut.
\newblock Conceptual captions: A cleaned, hypernymed, image alt-text dataset for automatic image captioning.
\newblock In \emph{Proceedings of the 56th Annual Meeting of the Association for Computational Linguistics (Volume 1: Long Papers)}, pages 2556--2565, 2018.

\bibitem[Sidorov et~al.(2020)Sidorov, Hu, Rohrbach, and Singh]{sidorov2020textcaps}
Oleksii Sidorov, Ronghang Hu, Marcus Rohrbach, and Amanpreet Singh.
\newblock Textcaps: a dataset for image captioning with reading comprehension.
\newblock In \emph{Computer Vision--ECCV 2020: 16th European Conference, Glasgow, UK, August 23--28, 2020, Proceedings, Part II 16}, pages 742--758. Springer, 2020.

\bibitem[Team(2023)]{team2023internlm}
InternLM Team.
\newblock Internlm: A multilingual language model with progressively enhanced capabilities, 2023.

\bibitem[Touvron et~al.(2023)Touvron, Martin, Stone, Albert, Almahairi, Babaei, Bashlykov, Batra, Bhargava, Bhosale, et~al.]{touvron2023llama}
Hugo Touvron, Louis Martin, Kevin Stone, Peter Albert, Amjad Almahairi, Yasmine Babaei, Nikolay Bashlykov, Soumya Batra, Prajjwal Bhargava, Shruti Bhosale, et~al.
\newblock Llama 2: Open foundation and fine-tuned chat models.
\newblock \emph{arXiv preprint arXiv:2307.09288}, 2023.

\bibitem[Vaswani et~al.(2017)Vaswani, Shazeer, Parmar, Uszkoreit, Jones, Gomez, Kaiser, and Polosukhin]{vaswani2017attention}
Ashish Vaswani, Noam Shazeer, Niki Parmar, Jakob Uszkoreit, Llion Jones, Aidan~N Gomez, {\L}ukasz Kaiser, and Illia Polosukhin.
\newblock Attention is all you need.
\newblock \emph{Advances in neural information processing systems}, 30, 2017.

\bibitem[Wu et~al.(2023)Wu, Zhang, Zhang, Chen, Liao, Wang, Li, Sun, Yan, Zhai, et~al.]{wu2023q}
Haoning Wu, Zicheng Zhang, Erli Zhang, Chaofeng Chen, Liang Liao, Annan Wang, Chunyi Li, Wenxiu Sun, Qiong Yan, Guangtao Zhai, et~al.
\newblock Q-bench: A benchmark for general-purpose foundation models on low-level vision.
\newblock \emph{arXiv preprint arXiv:2309.14181}, 2023.

\bibitem[Yang et~al.(2023)Yang, Xiao, Wang, Zhang, Yin, Lv, Pan, Wang, Yan, Yang, et~al.]{yang2023baichuan}
Aiyuan Yang, Bin Xiao, Bingning Wang, Borong Zhang, Chao Yin, Chenxu Lv, Da Pan, Dian Wang, Dong Yan, Fan Yang, et~al.
\newblock Baichuan 2: Open large-scale language models.
\newblock \emph{arXiv preprint arXiv:2309.10305}, 2023.

\bibitem[Ye et~al.(2023)Ye, Xu, Xu, Ye, Yan, Zhou, Wang, Hu, Shi, Shi, et~al.]{ye2023mplug}
Qinghao Ye, Haiyang Xu, Guohai Xu, Jiabo Ye, Ming Yan, Yiyang Zhou, Junyang Wang, Anwen Hu, Pengcheng Shi, Yaya Shi, et~al.
\newblock mplug-owl: Modularization empowers large language models with multimodality.
\newblock \emph{arXiv preprint arXiv:2304.14178}, 2023.

\bibitem[Yu et~al.(2023)Yu, Yang, Li, Wang, Lin, Liu, Wang, and Wang]{yu2023mm}
Weihao Yu, Zhengyuan Yang, Linjie Li, Jianfeng Wang, Kevin Lin, Zicheng Liu, Xinchao Wang, and Lijuan Wang.
\newblock Mm-vet: Evaluating large multimodal models for integrated capabilities.
\newblock \emph{arXiv preprint arXiv:2308.02490}, 2023.

\bibitem[Zhang et~al.(2022)Zhang, Zhang, Hu, Chen, Li, Dai, Wang, Yuan, Hwang, and Gao]{zhang2022glipv2}
Haotian Zhang, Pengchuan Zhang, Xiaowei Hu, Yen-Chun Chen, Liunian Li, Xiyang Dai, Lijuan Wang, Lu Yuan, Jenq-Neng Hwang, and Jianfeng Gao.
\newblock Glipv2: Unifying localization and vision-language understanding.
\newblock \emph{Advances in Neural Information Processing Systems}, 35:\penalty0 36067--36080, 2022.

\bibitem[Zhang et~al.(2023{\natexlab{a}})Zhang, Wang, Cao, Xu, Ouyang, Zhao, Ding, Zhang, Duan, Yan, et~al.]{zhang2023internlm}
Pan Zhang, Xiaoyi Dong~Bin Wang, Yuhang Cao, Chao Xu, Linke Ouyang, Zhiyuan Zhao, Shuangrui Ding, Songyang Zhang, Haodong Duan, Hang Yan, et~al.
\newblock Internlm-xcomposer: A vision-language large model for advanced text-image comprehension and composition.
\newblock \emph{arXiv preprint arXiv:2309.15112}, 2023{\natexlab{a}}.

\bibitem[Zhang et~al.(2023{\natexlab{b}})Zhang, Han, Zhou, Hu, Yan, Lu, Li, Gao, and Qiao]{zhang2023llama}
Renrui Zhang, Jiaming Han, Aojun Zhou, Xiangfei Hu, Shilin Yan, Pan Lu, Hongsheng Li, Peng Gao, and Yu Qiao.
\newblock Llama-adapter: Efficient fine-tuning of language models with zero-init attention.
\newblock \emph{arXiv preprint arXiv:2303.16199}, 2023{\natexlab{b}}.

\bibitem[Zhu et~al.(2023)Zhu, Chen, Shen, Li, and Elhoseiny]{zhu2023minigpt}
Deyao Zhu, Jun Chen, Xiaoqian Shen, Xiang Li, and Mohamed Elhoseiny.
\newblock Minigpt-4: Enhancing vision-language understanding with advanced large language models.
\newblock \emph{arXiv preprint arXiv:2304.10592}, 2023.

\end{thebibliography}
}

\end{document}